\documentclass{article}

\PassOptionsToPackage{numbers, compress}{natbib}
\usepackage[preprint]{neurips_2026}

\usepackage[utf8]{inputenc} 
\usepackage[T1]{fontenc}    
\usepackage{hyperref}       
\usepackage{url}            
\usepackage{booktabs}       
\usepackage{graphicx}       
\usepackage{amsmath}        
\usepackage{amsfonts}       
\usepackage{nicefrac}       
\usepackage{microtype}      
\usepackage{xcolor}         
\usepackage{array}          
\usepackage{colortbl}       
\usepackage{pifont}         
\usepackage{multirow}       

\newcommand{\method}{ViDAL}
\newcommand{\gain}[1]{\textcolor{green!45!black}{\scriptsize (#1)}}
\newcommand{\cmark}{\ding{51}}

\title{\method{}: A Visual Dynamics-Grounded Action Latent Space for Vision-Language-Action Models}

\author{%
  Yuan Xu\textsuperscript{1,2}\thanks{\texttt{yuan.xu@nlpr.ia.ac.cn}}\quad
  Yixiang Chen\textsuperscript{1,2}\quad
  Qisen Ma\textsuperscript{1,2}\quad
  Jiabing Yang\textsuperscript{1,2}\quad
  Peiyan Li\textsuperscript{1,2} \\[2pt]
  \bfseries
  Kai Wang\textsuperscript{1,2}\quad
  Jianhua Yang\textsuperscript{1,2}\quad
  Jianlou Si\textsuperscript{3}\quad
  Jun Huang\textsuperscript{3} \\[2pt]
  \bfseries
  Jing Liu\textsuperscript{1,2,4}\quad
  Nianfeng Liu\textsuperscript{1,2,4}\quad
  Yan Huang\textsuperscript{1,2,4}\quad
  Liang Wang\textsuperscript{1,2} \\[6pt]
  \normalfont
  \textsuperscript{1}CASIA \quad
  \textsuperscript{2}UCAS \quad
  \textsuperscript{3}Alibaba Group \quad
  \textsuperscript{4}FiveAges%
}

\begin{document}

\maketitle

\begin{abstract}
Vision-Language-Action (VLA) models have become a central paradigm for robot policy learning, which predict actions in three forms: raw action chunks, discrete action tokens, or continuous action latents. However, existing action representations primarily model action trajectories, with limited consideration of the visual dynamics induced by these actions. We introduce \method{}, a Visual Dynamics-grounded Action Latent Space that anchors continuous action latents in the future visual dynamics of the scene. Specifically, \method{} learns action latent space by training an Action Variational Autoencoder (Action VAE) to reconstruct action chunks while aligning its latent with future scene dynamics. When integrated into downstream robot policies, the proposed Action VAE serves as a plug-in action interface compatible with multiple VLA architectures and enables optional future-video prediction as an additional capability. Empirically, \method{} outperforms competitive baselines on LIBERO with 98.1\% average success, improves a multi-task $\pi_{0.5}$ policy on RoboTwin~2.0 from 54.3\% to 65.5\% (Clean) and from 33.2\% to 43.1\% (Random) success rates over 50 dual-arm tasks, and yields 20.0\% and 23.4\% absolute success-rate gains on real-world single-arm Franka and dual-arm ARX robot platforms.
\end{abstract}

\section{Introduction}

Vision-Language-Action (VLA) models~\citep{brohan2022rt1,zitkovich2023rt2,kim2024openvla,black2024pi0,intelligence2025pi05,liu2024rdt} have emerged as a popular paradigm for generalist robot manipulation, directly predicting actions from visual observations and language instructions. Since actions define the interface between the VLA policy and robot control, their representation determines both the learning target and the executable control output. In recent VLA systems, action representations usually take three predominant forms: raw action chunks regressed by a continuous head~\citep{black2024pi0,intelligence2025pi05,chi2023diffusion}, discrete action tokens produced by a tokenizer~\citep{pertsch2025fast,wang2025vqvla,zhou2025beast,faster2025}, or continuous action latents produced by a learned autoencoder~\citep{tan2024rold}. These representations primarily model action trajectories, but manipulation actions are meaningful through their effects on the scene, such as object motion, contact changes, and task-state transitions. Accordingly, as VLA gradually scales to long-horizon, multi-task and multi-scenario manipulation~\citep{walke2023bridgedata,openx2023,khazatsky2024droid,octo2024}, an effective action representation should account for not only the action trajectory but also the visual dynamics induced by the action.

\begin{figure}[t]
  \centering
  \includegraphics[width=\linewidth]{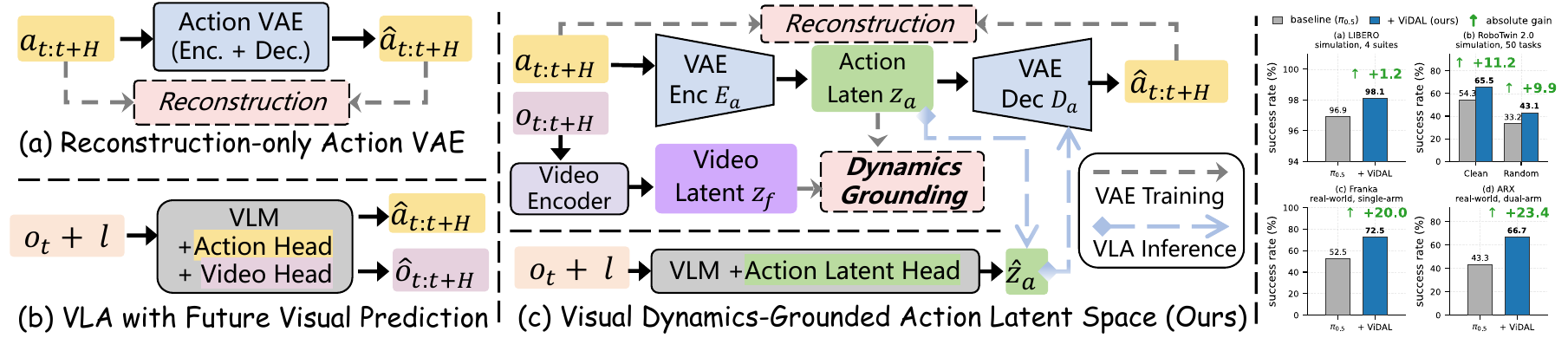}
  \caption{\textbf{Overview of \method{}.} (a) Action VAE supervised only by reconstruction. (b) VLAs supervised with visual prediction signals. (c) \method{} (ours) grounds the action latent space in future visual dynamics, and uses the Action VAE as a plug-in action representation at deployment. Right column: consistent success-rate gains on simulation benchmarks and realrobot platforms.}
  \label{fig:teaser}
\end{figure}

Recent works have pursued two complementary directions: learning compact action representations and incorporating visual dynamics into VLA policies. As shown in Figure~\ref{fig:teaser}(a), early action tokenizers or autoencoders compress action chunks into compact tokens or latents~\citep{pertsch2025fast,wang2025vqvla,zhou2025beast,faster2025,tan2024rold}, yielding representations primarily optimized for action reconstruction, without modeling visual dynamics. As shown in Figure~\ref{fig:teaser}(b), VLAs with visual prediction use future video or world-model signals to guide policy learning~\citep{assran2025vjepa2,hu2025vpp,cen2025worldvla,xu2026futurevla,shang2026dynvla,su2026worldguidance}. Together, these directions suggest that visual dynamics benefit action learning, but leave open how they should shape the action representation. Since actions are visually contextual, the challenge is learning an executable, visual dynamics-grounded action latent space.

We propose \method{} (Visual Dynamics-grounded Action Latent Space, Figure~\ref{fig:teaser}(c)) to inject visual dynamics directly into the action latent during representation learning. In \method{}, a frozen video VAE extracts the initial and future visual latents from each robot trajectory, and an Action VAE encodes the corresponding action chunk into a continuous action latent. A lightweight visual-dynamics predictor then maps the initial visual latent and the action latent to the future visual latent, training the action latent to preserve executable actions while explaining future scene dynamics conditioned on the current observation. For downstream policy learning, only the frozen Action VAE is required: its encoder produces action latent targets for VLA training, and its decoder maps VLA-predicted latents to executable robot actions. Thus, the decoupled design lets \method{} plug into both regression (e.g., OpenVLA-OFT~\citep{kim2025openvlaoft}) and flow-matching (e.g., $\pi_{0.5}$~\citep{intelligence2025pi05}) action experts without modifying the VLM backbone. The video VAE also enables future-video prediction for visualizing the learned dynamics.

We evaluate \method{} on two simulation benchmarks (LIBERO~\citep{liu2024libero}, RoboTwin~2.0~\citep{mu2024robotwin}) and two real-world platforms (single-arm Franka, dual-arm ARX), using both regression (OpenVLA-OFT~\citep{kim2025openvlaoft}, StarVLA-OFT~\citep{ye2026starvlaalpha}) and flow-matching ($\pi_{0.5}$~\citep{intelligence2025pi05}) action experts as the downstream policy. Across all four settings, \method{} consistently improves the underlying VLA policies and surpasses competitive action-representation baselines on LIBERO (Figure~\ref{fig:teaser}, right): on the higher-dimensional dual-arm RoboTwin~2.0, it lifts a multi-task $\pi_{0.5}$ policy by $11.2\%$ Clean and $9.9\%$ Random over 50 tasks, and on real robots it delivers $20.0\%$ absolute gain on Franka and $23.4\%$ on ARX. These significant gains across simulation and real, single- and dual-arm settings establish visual-dynamics grounding as a better design principle for action representations.

Our contributions are summarized as follows:
\begin{itemize}
  \item We introduce \method{}, which, to our knowledge, is the first approach to ground action latents in future visual dynamics for VLA models, learned via an Action VAE jointly trained on action reconstruction and future visual latent prediction.
  \item We propose a plug-in action-latent interface for downstream VLA policies, where the frozen Action VAE serves as the action representation for both regression and flow-matching action experts, requiring no changes to the VLM backbone.
  \item We conduct extensive experiments in both simulation (LIBERO, RoboTwin~2.0) and real-world (single-arm Franka, dual-arm ARX) environments. Results show that \method{} consistently improves both regression and flow-matching VLA policies and surpasses competitive action-representation methods.
\end{itemize}

\section{Related Work}

\noindent\textbf{Vision-Language-Action Models.}
Vision-Language-Action (VLA) models adapt pretrained vision-language backbones to predict robot actions from visual observations and language instructions~\citep{brohan2022rt1,zitkovich2023rt2,kim2024openvla,driess2023palme,li2024roboflamingo,octo2024,liu2024rdt}. Recent VLAs differ mainly in how their action experts generate actions: flow-matching experts in $\pi_0$~\citep{black2024pi0} and $\pi_{0.5}$~\citep{intelligence2025pi05} support high-frequency continuous control, while one-step regression experts in OpenVLA-OFT~\citep{kim2025openvlaoft} and StarVLA-OFT~\citep{ye2026starvlaalpha} target faster inference. \method{} is orthogonal to these backbones and experts: it changes the action target to a dynamics-grounded latent and decodes the predicted latent back to executable actions, without modifying the underlying VLA architecture.

\noindent\textbf{Action Representations for VLA Models.}
VLA action representations take three predominant forms, each shaped by a different supervision signal. Raw action chunks are regressed under behavior-cloning~\citep{black2024pi0,intelligence2025pi05,liu2024rdt,chi2023diffusion,zhao2023act}; discrete action tokens compress trajectories via quantization~\citep{pertsch2025fast,wang2025vqvla,zhou2025beast,faster2025}; and continuous action latents use an autoencoder bottleneck without quantization~\citep{tan2024rold}. All three are shaped only by reconstruction, token efficiency, or imitation losses, and rarely consider how the scene visually evolves under each action. \method{} explicitly introduces visual dynamics as a direct supervision signal for the action latent, anchoring the representation to the visual effect of each action chunk.

\noindent\textbf{Visual Dynamics for VLA Policies.}
There is also a complementary line that uses visual dynamics to guide VLA policies. JEPA-style methods pretrain on video to provide dynamics-aware features that condition the policy~\citep{assran2025vjepa2,sun2026vlajepa}. Future-prediction approaches inject future-frame or video-diffusion signals into the VLM backbone or the policy as auxiliary supervision~\citep{hu2025vpp,jia2025video2act,dreamzero2026}. Other methods use dynamics or world-model signals for action reasoning~\citep{cen2025worldvla,shang2026dynvla,xu2026futurevla,su2026worldguidance}. These designs entangle dynamics learning with policy training and tie the visual signal to a VLA architecture, so the dynamics machinery must be retrained for every new backbone. \method{} confines visual dynamics to the action latent during representation learning and freezes it before policy training, so the same dynamics-grounded Action VAE can be plugged into different VLM backbones and action experts.

\section{Method}

\subsection{Overview}

\begin{figure}[t]
  \centering
  \includegraphics[width=0.98\linewidth]{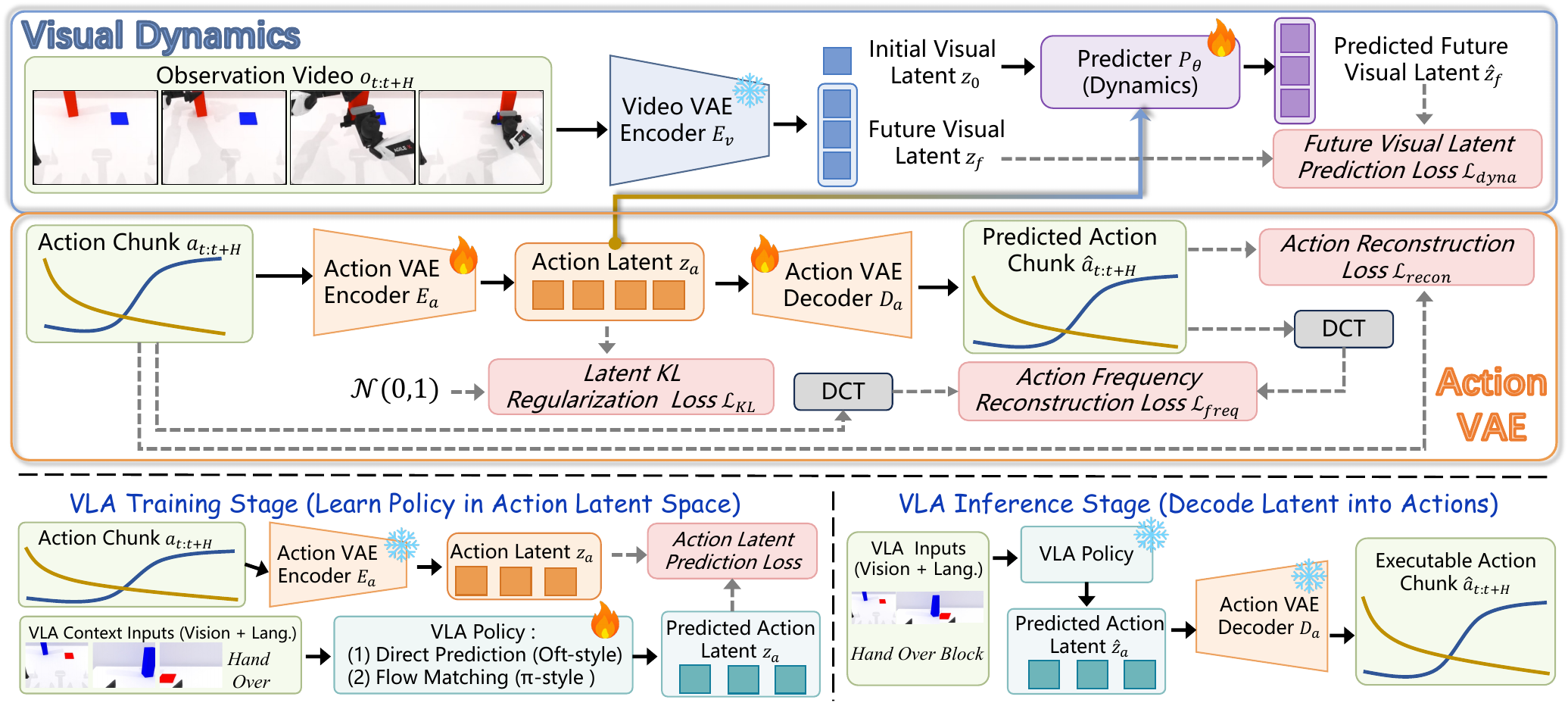}
  \caption{\textbf{Framework of \method{}.} \textit{Top:} the Action VAE is trained with visual-dynamics supervision from a frozen video encoder. \textit{Bottom:} the frozen Action VAE then plugs into VLA training (left) and inference (right), serving as the action representation between the policy and the robot.}
  \label{fig:method_overview}
\end{figure}

\method{} learns a reusable action representation whose continuous latent is grounded by visual dynamics. As shown in Figure~\ref{fig:method_overview}, given a robot trajectory with observation clip $\mathbf{o}_{t:t+H}$ and action chunk $\mathbf{a}_{t:t+H-1}$, an Action VAE encodes the action chunk into a continuous action latent that is jointly supervised by action reconstruction and visual-dynamics prediction based on video VAE model(Section~\ref{sec:method_visual_dynamics}). Then the frozen Action VAE serves as a plug-in action representation for downstream VLA policy learning (Section~\ref{sec:method_vla_policy}), where the video model is no longer required.


\subsection{Learning the Action Latent Space}
\label{sec:method_visual_dynamics}


The central challenge in learning action latents is that action reconstruction alone does not determine which aspects of an action chunk are relevant to policy learning: a low-dimensional latent may reconstruct actions while failing to encode how those actions interact with the current scene. \method{} addresses this by adding visual-dynamics supervision, encouraging the action latent to be informative for predicting the future visual latent conditioned on the initial observation.

\noindent\textbf{Action VAE.}
The Action VAE maps the action chunk to a continuous action latent $\mathbf{z}_a\in\mathbb{R}^{T\times D}$ via a Gaussian posterior $q_\phi(\mathbf{z}_a\!\mid\!\mathbf{a}_{t:t+H-1}) = \mathcal{N}(\boldsymbol{\mu}_\phi,\,\mathrm{diag}(\boldsymbol{\sigma}_\phi^2))$, where the encoder applies a $4{\times}$ temporal downsampling so that $T{=}H/4$ and $D$ is the latent dimension. We sample $\mathbf{z}_a$ with the standard reparameterization trick and decode it back to the original action chunk:
\begin{equation}
  \mathbf{z}_a = \boldsymbol{\mu}_\phi + \boldsymbol{\sigma}_\phi \odot \boldsymbol{\epsilon},\quad
  \boldsymbol{\epsilon}\sim\mathcal{N}(\mathbf{0},\mathbf{I}),\qquad
  \hat{\mathbf{a}}_{t:t+H-1} = D_a^\psi(\mathbf{z}_a).
  \label{eq:action_vae}
\end{equation}
The continuous bottleneck preserves smooth trajectory structure and is compatible with both regression and flow-matching action experts.

\noindent\textbf{Visual-Dynamics Supervision.}
A frozen video encoder $E_v$ extracts an initial visual latent $\mathbf{z}_0$ from the current observation and a future visual latent $\mathbf{z}_f$ from the subsequent frames as the prediction target,
\begin{equation}
  (\mathbf{z}_0,\mathbf{z}_f) = E_v(\mathbf{o}_{t:t+H}),
  \label{eq:video_latents}
\end{equation}
and a lightweight predictor $P_\theta$ regresses $\mathbf{z}_f$ from the initial visual latent and the action latent:
\begin{equation}
  \hat{\mathbf{z}}_f = P_\theta(\mathbf{z}_0, \mathbf{z}_a).
  \label{eq:future_prediction}
\end{equation}
Conditioning on $\mathbf{z}_0$ is essential because the same action chunk can correspond to different scene transitions in different contexts. The predictor therefore drives $\mathbf{z}_a$ to retain action-relevant scene-transition structure rather than only low-level action statistics. Both $E_v$ and $P_\theta$ are used only during representation learning and discarded afterwards, so deployment requires no video model.

\noindent\textbf{Training Objective.}
The Action VAE is jointly trained with four complementary terms,
\begin{equation}
  \mathcal{L}_{\text{\method{}}}
  = \mathcal{L}_{\mathrm{rec}}
  + \lambda_{\mathrm{freq}}\,\mathcal{L}_{\mathrm{freq}}
  + \lambda_{\mathrm{KL}}\,\mathcal{L}_{\mathrm{KL}}
  + \lambda_{\mathrm{dyna}}\,\mathcal{L}_{\mathrm{dyna}},
  \label{eq:method_objective}
\end{equation}
where $\mathcal{L}_{\mathrm{rec}}\!=\!\|\hat{\mathbf{a}}-\mathbf{a}\|_1$ reconstructs the action chunk in the time domain, $\mathcal{L}_{\mathrm{freq}}\!=\!\|\mathrm{DCT}(\hat{\mathbf{a}})-\mathrm{DCT}(\mathbf{a})\|_1$ follows FAST's DCT-based action tokenization~\citep{pertsch2025fast} to preserve temporal frequency structure, $\mathcal{L}_{\mathrm{KL}}\!=\!D_{\mathrm{KL}}(q_\phi(\mathbf{z}_a\!\mid\!\mathbf{a})\,\|\,\mathcal{N}(\mathbf{0},\mathbf{I}))$ regularizes the latent distribution, and $\mathcal{L}_{\mathrm{dyna}}\!=\!\|\hat{\mathbf{z}}_f-\mathrm{sg}(\mathbf{z}_f)\|_1$ aligns the action latent with future scene evolution; the stop-gradient operator $\mathrm{sg}(\cdot)$ keeps the visual target space fixed.

\subsection{Integrating \method{} into VLA Policies}
\label{sec:method_vla_policy}

After representation learning, \method{} replaces the raw-action interface of a VLA policy with an interface based on action latents: during training, the policy predicts an encoded action latent, and at inference, the predicted latent is decoded into an executable action chunk.

\noindent\textbf{Plug-In Action Representation.}
The Action VAE is frozen and used as a plug-in action representation that the VLA policy is trained to predict. For each action chunk $\mathbf{a}_{t:t+H-1}$, the frozen encoder produces a deterministic latent target,
\begin{equation}
  \bar{\mathbf{z}}_a = \mathrm{sg}\!\left(\mathrm{Norm}\!\left(\boldsymbol{\mu}_\phi(\mathbf{a}_{t:t+H-1})\right)\right),
  \label{eq:latent_action_target}
\end{equation}
where $\mathrm{Norm}(\cdot)$ is a fixed latent normalization estimated from the Action VAE training set, and the bar distinguishes this deterministic downstream target from the sampled latent $\mathbf{z}_a$ used in Action VAE training. This interface keeps the VLM backbone unchanged and only replaces the action-modeling target, so the same dynamics-grounded latent space can be plugged into different VLA architectures without modifying their backbones.

\noindent\textbf{Regression Action Experts.}
For VLAs with regression action experts such as OpenVLA-OFT~\citep{kim2025openvlaoft} and StarVLA-OFT~\citep{ye2026starvlaalpha}, the action expert maps the VLA context $\mathbf{c}$ (visual observation, language instruction, and policy features) to a predicted latent $\tilde{\mathbf{z}}_a$ and is trained against $\bar{\mathbf{z}}_a$:
\begin{equation}
  \tilde{\mathbf{z}}_a = F_\omega(\mathbf{c}),\qquad
  \mathcal{L}_{\mathrm{reg}} = \ell\!\left(\tilde{\mathbf{z}}_a,\bar{\mathbf{z}}_a\right),
  \label{eq:reg_objective}
\end{equation}
where $\ell(\cdot,\cdot)$ is the regression loss used by the action expert. The lightweight one-step expert is preserved while the supervised target moves from the raw action space to the dynamics-grounded latent space.

\noindent\textbf{Flow-Matching Action Experts.}
For VLAs with flow-matching action experts such as $\pi_0$ and $\pi_{0.5}$~\citep{black2024pi0,intelligence2025pi05}, the same latent target serves as the endpoint of a conditional flow. With the interpolant $\bar{\mathbf{z}}_\tau = \tau\bar{\mathbf{z}}_a + (1-\tau)\boldsymbol{\xi}$ and noise $\boldsymbol{\xi}\sim\mathcal{N}(\mathbf{0},\mathbf{I})$, the action expert learns a vector field in the action latent space:
\begin{equation}
  \mathcal{L}_{\mathrm{flow}}
  = \mathbb{E}_{\tau,\boldsymbol{\xi},\bar{\mathbf{z}}_a}
  \!\left[\,\big\| v_\omega(\bar{\mathbf{z}}_\tau,\mathbf{c},\tau) - (\boldsymbol{\xi}-\bar{\mathbf{z}}_a)\big\|_2^2\,\right].
  \label{eq:flow_objective}
\end{equation}
The generative process of flow-matching policies is unchanged, and only the modeled distribution shifts from raw actions to dynamics-grounded action latents.

\noindent\textbf{Inference.}
The VLA policy emits a predicted latent $\tilde{\mathbf{z}}_a$ and the frozen Action VAE decoder maps it back to executable actions, $\hat{\mathbf{a}}_{t:t+H-1} = D_a^\psi(\tilde{\mathbf{z}}_a)$. \method{} therefore acts as a drop-in action representation layer between any existing VLA action expert and the robot, with no video model in the loop.

\section{Experiments}

We evaluate \method{} as a plug-in action representation for VLA policies, demonstrating whether visual-dynamics grounding improves downstream success across simulation and real robots. Following the experimental setup (Section~\ref{sec:exp_setup}), the experiments are organized around three questions:
\begin{enumerate}
  \item Does \method{} consistently improve VLA policies on simulation benchmarks (Section~\ref{sec:exp_sim}), and do the gains transfer to real-robot platforms (Section~\ref{sec:exp_real_world})?
  \item Which training-objective terms and action-latent design choices drive these improvements (Section~\ref{sec:exp_ablation}), and how do they scale with tasks and demonstrations (Section~\ref{sec:exp_further_analysis})?
  \item Are \method{} action latents more structured than raw actions, and do they correspond to recognizable future visual dynamics (Section~\ref{sec:exp_qualitative})?
\end{enumerate}

\subsection{Experimental Setup}
\label{sec:exp_setup}

\noindent\textbf{Benchmarks.}
We evaluate on two simulation benchmarks and two real-world platforms. LIBERO~\citep{liu2024libero} covers single-arm language-conditioned manipulation across four task suites (Object, Spatial, Goal, Long); RoboTwin~2.0~\citep{mu2024robotwin} covers dual-arm manipulation across 50 tasks under a Clean setting matched to training and a Random setting with randomized scene conditions (e.g., background, lighting, distractors, viewpoint). Real-world deployment uses a single-arm Franka and a dual-arm ARX, each with a small set of tabletop tasks; we report task success rates (\%) throughout.

\noindent\textbf{VLA backbones and baselines.}
We pair \method{} with both regression and flow-matching action experts: OpenVLA-OFT~\citep{kim2025openvlaoft} on LIBERO and StarVLA-OFT~\citep{ye2026starvlaalpha} on RoboTwin~2.0 for regression, and $\pi_{0.5}$~\citep{intelligence2025pi05} on both benchmarks for flow matching; the primary comparison contrasts raw-action targets against \method{} latent targets on each backbone. On LIBERO we additionally compare against action-tokenization VLAs ($\pi_0$-FAST~\citep{pertsch2025fast}, FASTer~\citep{faster2025}) and recent VLAs that augment the VLM backbone with visual dynamics or world models~\citep{bjorck2025gr00t,cen2025worldvla,zhang2025dreamvla,bu2025univla,chen2025villa}.

\noindent\textbf{Training and evaluation.}
We train \method{} per benchmark on its demonstrations with a frozen Wan~VAE~\citep{wan2025wan} as the video encoder, and reuse it as a frozen plug-in action representation across all VLA backbones. The latent dimension is $D{=}16$ for the dual-arm settings (RoboTwin~2.0, ARX) and $D{=}8$ for the single-arm settings (LIBERO, Franka), with loss weights $\lambda_{\mathrm{dyna}}{=}1$ and $\lambda_{\mathrm{KL}}{=}10^{-4}$ throughout. Each VLA policy is trained as a multi-task policy across all tasks of its benchmark, following the backbone's original recipe with the action target replaced by the \method{} latent.

\subsection{Main Results}
\label{sec:exp_sim}

\noindent\textbf{LIBERO.}
On LIBERO, \method{} yields absolute success-rate gains of $+0.3\%$ on OpenVLA-OFT and $+1.2\%$ on $\pi_{0.5}$, concentrated on the long-horizon Long suite (Table~\ref{tab:libero_main}). The modest overall margin reflects LIBERO's near-saturation regime, whereas the Long-suite gain supports our claim that visual-dynamics grounding is most beneficial under extended temporal horizons. $\pi_{0.5}$+\method{} still attains the highest average across the comparison, on par with the strongest action-tokenization and recent VLA baselines without action discretization or auxiliary supervision on the VLM backbone.

\noindent\textbf{RoboTwin 2.0.}
On the dual-arm RoboTwin~2.0, \method{} yields absolute gains of $+11.2\%$/$+9.9\%$ on $\pi_{0.5}$ and $+4.2\%$/$+1.8\%$ on StarVLA-OFT under the Clean/Random splits (Table~\ref{tab:robotwin_main}). Higher-DoF coordination and longer horizons stress the action representation more than near-saturation LIBERO, and the larger flow-matching gain parallels its distributional action head, which couples more naturally with a continuous dynamics-grounded latent than the one-step regression head of StarVLA-OFT. The gain further persists under Random scene perturbations, allowing the multi-task $\pi_{0.5}$+\method{} to surpass the official single-task $\pi_0$ specialists on both splits.

\begin{table}[t]
  \caption{Success rates (\%) on LIBERO. Methods are grouped; gray rows are our +\method{} variants and deltas are relative to the corresponding backbone.}
  \label{tab:libero_main}
  \centering
  \resizebox{0.7\linewidth}{!}{
  \begin{tabular}{lccccc}
    \toprule
    Method & Object & Spatial & Goal & Long & Average \\
    \midrule
    \multicolumn{6}{l}{\textit{Action-tokenization VLAs}} \\
    $\pi_0$-FAST~\citep{pertsch2025fast} & 97.2 & 96.6 & 96.0 & 86.8 & 94.2 \\
    FASTer~\citep{faster2025} & \textbf{99.4} & 98.0 & \textbf{98.6} & 95.4 & 97.9 \\
    \midrule
    \multicolumn{6}{l}{\textit{Visual-Dynamics Augmented VLAs}} \\
    GR00T-N1.5~\citep{bjorck2025gr00t} & 97.6 & 94.4 & 93.0 & 90.6 & 93.9 \\
    WorldVLA~\citep{cen2025worldvla} & 96.2 & 87.6 & 83.4 & 60.0 & 81.8 \\
    DreamVLA~\citep{zhang2025dreamvla} & 94.0 & 97.5 & 89.5 & 89.5 & 92.6 \\
    UniVLA~\citep{bu2025univla} & 96.8 & 96.5 & 95.6 & 92.0 & 95.2 \\
    Villa-X~\citep{chen2025villa} & 97.0 & 97.5 & 91.5 & 74.5 & 90.1 \\
    \midrule
    \multicolumn{6}{l}{\textit{VLA backbones (vanilla vs.\ +\method{})}} \\
    OpenVLA-OFT~\citep{kim2025openvlaoft}(1 policy for all 4 suites) & 98.0 & 97.7 & 96.1 & 95.3 & 96.8 \\
    \rowcolor{gray!10} OpenVLA-OFT+\method{} & 97.8 & 98.6 & 96.2 & \textbf{95.6} & 97.1\gain{+0.3} \\
    $\pi_{0.5}$~\citep{intelligence2025pi05} & 98.2 & 98.8 & 98.0 & 92.4 & 96.9 \\
    \rowcolor{gray!10} $\pi_{0.5}$+\method{} & \textbf{99.4} & \textbf{99.2} & 98.2 & \textbf{95.6} & \textbf{98.1}\gain{+1.2} \\
    \bottomrule
  \end{tabular}}
\end{table}

\begin{table}[t]
  \caption{Success rates (\%) on RoboTwin 2.0 under Clean/Random settings; we show 15 representative tasks and report Average over all 50 tasks. Gray columns are our +\method{} variants and deltas are relative to the corresponding backbone.}
  \label{tab:robotwin_main}
  \centering
  \resizebox{0.98\textwidth}{!}{
  \begin{tabular}{lrr|rrrrrrrr}
    \toprule
    & \multicolumn{2}{c|}{\makebox[0.14\textwidth][c]{$\pi_0$~{\scriptsize\citep{black2024pi0}}}} &
    \multicolumn{2}{c}{\makebox[0.14\textwidth][c]{StarVLA-OFT~{\scriptsize\citep{ye2026starvlaalpha}}}} &
    \multicolumn{2}{c}{\cellcolor{gray!10}\makebox[0.14\textwidth][c]{\shortstack{StarVLA-OFT\\+\method{}}}} &
    \multicolumn{2}{c}{\makebox[0.14\textwidth][c]{$\pi_{0.5}$~{\scriptsize\citep{intelligence2025pi05}}}} &
    \multicolumn{2}{c}{\cellcolor{gray!10}\makebox[0.14\textwidth][c]{\shortstack{$\pi_{0.5}$\\+\method{}}}} \\
    \cmidrule(lr){2-3} \cmidrule(lr){4-5} \cmidrule(lr){6-7} \cmidrule(lr){8-9} \cmidrule(lr){10-11}
    Task & Clean & Random & Clean & Random & Clean & Random & Clean & Random & Clean & Random \\
    \midrule
    Adjust Bottle & 90.0 & 56.0 & 96.0 & 0.0 & \textbf{100.0} & 5.0 & 84.0 & 73.0 & 99.0 & \textbf{86.0} \\
    Place Empty Cup & 37.0 & 11.0 & 72.0 & 4.0 & 66.0 & 0.0 & 80.0 & 56.0 & \textbf{89.0} & \textbf{83.0} \\
    Click Alarmclock & 63.0 & 11.0 & \textbf{91.0} & 14.0 & 81.0 & 6.0 & 56.0 & 42.0 & 79.0 & \textbf{79.0} \\
    Open Laptop & 85.0 & 46.0 & 31.0 & 0.0 & 79.0 & 6.0 & 75.0 & 48.0 & \textbf{91.0} & \textbf{77.0} \\
    Place Burger Fries & 80.0 & 4.0 & 96.0 & 6.0 & \textbf{97.0} & 4.0 & 68.0 & 47.0 & 85.0 & \textbf{77.0} \\
    Dump Bin Bigbin & 83.0 & 24.0 & 68.0 & 1.0 & 87.0 & 7.0 & 86.0 & 54.0 & \textbf{92.0} & \textbf{69.0} \\
    Move Playingcard Away & 53.0 & 22.0 & 69.0 & 1.0 & \textbf{93.0} & 5.0 & 64.0 & 35.0 & 73.0 & \textbf{68.0} \\
    \textit{$\cdots$ (50 tasks total)} &&&&&&&&&& \\
    Handover Mic & 98.0 & 13.0 & 39.0 & 0.0 & 90.0 & 1.0 & 75.0 & 26.0 & \textbf{100.0} & \textbf{54.0} \\
    Press Stapler & 62.0 & 29.0 & 60.0 & 0.0 & 78.0 & 11.0 & 66.0 & 50.0 & \textbf{83.0} & \textbf{55.0} \\
    Open Microwave & \textbf{80.0} & 50.0 & -- & 2.0 & 60.0 & 15.0 & 63.0 & 31.0 & 75.0 & \textbf{53.0} \\
    Stack Bowls Three & 66.0 & 24.0 & 62.0 & 0.0 & 32.0 & 0.0 & 53.0 & 44.0 & \textbf{78.0} & \textbf{52.0} \\
    Place Bread Basket & 17.0 & 4.0 & 52.0 & 5.0 & \textbf{61.0} & 2.0 & 41.0 & 33.0 & 57.0 & \textbf{46.0} \\
    Pick Dual Bottles & 57.0 & 12.0 & 43.0 & 13.0 & 63.0 & 8.0 & 54.0 & \textbf{49.0} & \textbf{69.0} & 42.0 \\
    Put Bottles Dustbin & 54.0 & 13.0 & -- & 0.0 & 9.0 & 1.0 & 47.0 & 36.0 & \textbf{68.0} & \textbf{42.0} \\
    Stack Bowls Two & 91.0 & 41.0 & 90.0 & 1.0 & 67.0 & 0.0 & 92.0 & 66.0 & \textbf{99.0} & \textbf{75.0} \\
    \midrule
    \textbf{Average} & 46.4 & 16.3 & 50.4 & 2.2 & \cellcolor{gray!10}54.6\gain{+4.2} & \cellcolor{gray!10}4.0\gain{+1.8} & 54.3 & 33.2 & \cellcolor{gray!10}\textbf{65.5}\gain{+11.2} & \cellcolor{gray!10}\textbf{43.1}\gain{+9.9} \\
    \bottomrule
  \end{tabular}}
\end{table}

\begin{table}[t]
  \caption{Training objective ablation of \method{} on RoboTwin~2.0 with $\pi_{0.5}$. Clean/Random success rates (\%) average over 50 tasks; reconstruction loss is always on.}
  \label{tab:robotwin_ablation}
  \centering
  \resizebox{0.55\linewidth}{!}{
  \begin{tabular}{lcccrr}
    \toprule
    Variant & $\mathcal{L}_{\mathrm{dyna}}$ & $\mathcal{L}_{\mathrm{freq}}$ & $\mathcal{L}_{\mathrm{KL}}$ & Clean & Random \\
    \midrule
    Vanilla Action VAE                &        &        & \cmark & 57.4 & 38.3 \\
    w/o $\mathcal{L}_{\mathrm{dyna}}$ &        & \cmark & \cmark & 62.2 & 40.1 \\
    w/o $\mathcal{L}_{\mathrm{freq}}$ & \cmark &        & \cmark & 63.1 & 40.4 \\
    w/o $\mathcal{L}_{\mathrm{KL}}$   & \cmark & \cmark &        & 63.2 & 39.7 \\
    \rowcolor{gray!10}
    Full \method{}                    & \cmark & \cmark & \cmark & \textbf{65.5} & \textbf{43.1} \\
    \bottomrule
  \end{tabular}}
\end{table}

\setcounter{topnumber}{3}    
\setcounter{totalnumber}{4}  
\renewcommand{\topfraction}{0.92} 

\subsection{Ablation Studies}
\label{sec:exp_ablation}

We ablate \method{} along two axes: the training objective and the four latent design choices ($D$, $H$, $\lambda_{\mathrm{dyna}}$, $\lambda_{\mathrm{KL}}$). All ablations retrain the Action VAE from scratch and use $\pi_{0.5}$ as the downstream policy on RoboTwin~2.0; we report Clean/Random average success rates over all 50 tasks.

\noindent\textbf{Training objective.}
Each loss term in $\mathcal{L}_{\text{\method{}}}$ complements the others (Table~\ref{tab:robotwin_ablation}). A vanilla Action VAE with only time-domain reconstruction and a KL prior provides a lower bound, and disabling either the DCT-domain reconstruction $\mathcal{L}_{\mathrm{freq}}$ or the visual-dynamics term $\mathcal{L}_{\mathrm{dyna}}$ degrades both splits. The drop is steepest for $\mathcal{L}_{\mathrm{dyna}}$, confirming that visual-dynamics supervision is the structural source of the gain while $\mathcal{L}_{\mathrm{freq}}$ adds a smaller frequency-domain refinement.

\begin{figure}[t]
  \centering
  \includegraphics[width=\linewidth]{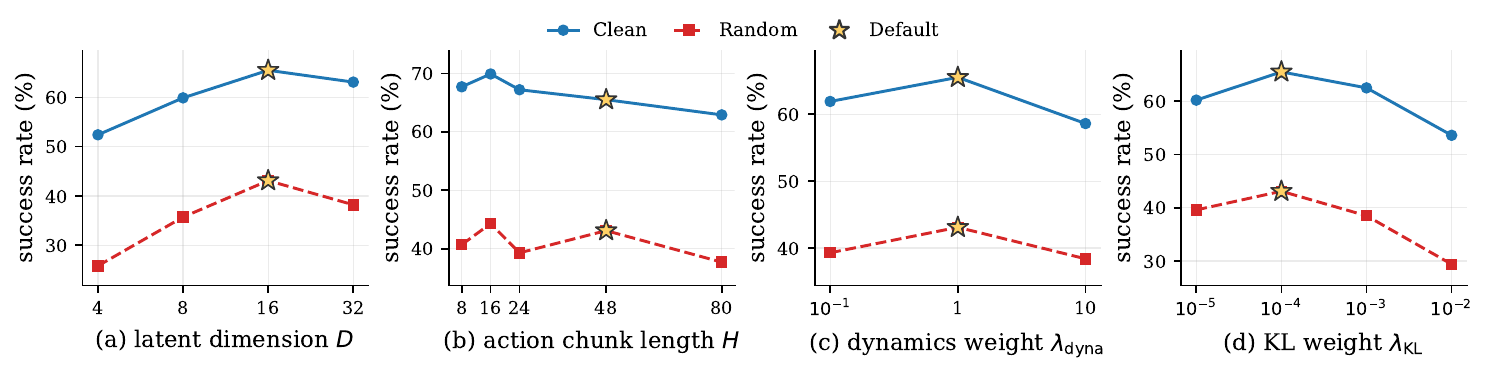}
  \caption{Action latent design ablation of \method{} on RoboTwin~2.0 with $\pi_{0.5}$. Each panel sweeps one design choice with the others fixed; stars mark the default.}
  \label{fig:latent_design}
\end{figure}

\noindent\textbf{Latent design choices.}
All four design choices show clear trade-offs around the default (Figure~\ref{fig:latent_design}). The latent dimension $D$ and chunk length $H$ follow inverted-U trends: too small a $D$ or short an $H$ underfits, while too large a $D$ or long an $H$ loosens the prior or compresses inconsistent sub-segments into one code. $\lambda_{\mathrm{dyna}}$ behaves similarly: weak supervision loses pixel-space anchoring, while overly strong values divert capacity from action reconstruction. $\lambda_{\mathrm{KL}}$ is the sharpest factor, since an over-strong prior collapses the latent and an over-weak prior leaves it under-regularized for the flow expert. The default thus balances reconstruction, dynamics anchoring, and prior regularization rather than maximizing any single axis.

\begin{figure}[t]
  \centering
  \includegraphics[width=\linewidth]{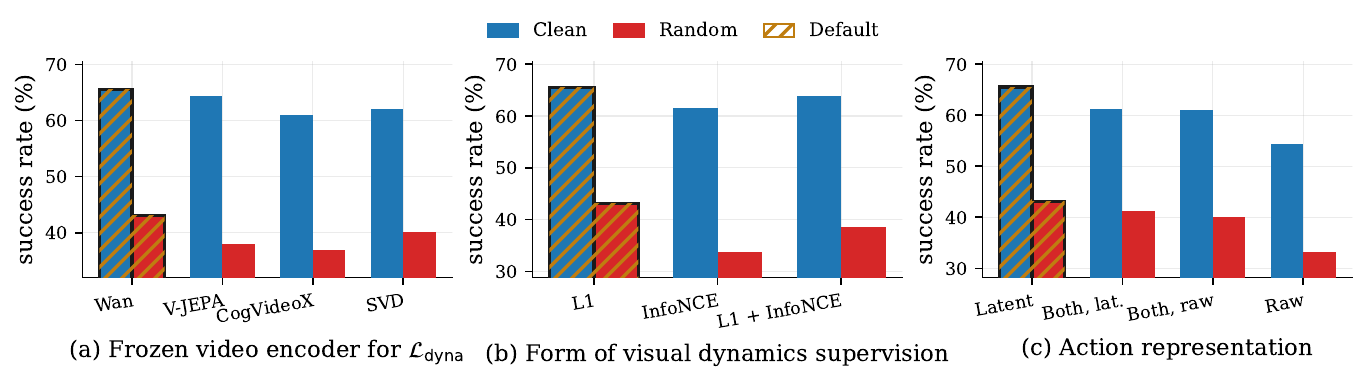}
  \caption{Component comparison for \method{} on RoboTwin~2.0 with $\pi_{0.5}$. (a)~frozen Video Encoder or Video VAE Encoder for $\mathcal{L}_{\mathrm{dyna}}$; (b)~form of dynamics supervision; (c)~policy prediction target and inference decoding branch (\textit{Latent}/\textit{Raw}: predict only the latent or only raw actions; \textit{Both,\,lat./raw}: predict both, decode from the named branch).}
  \label{fig:further_categorical}
\end{figure}

\subsection{Further Analysis}
\label{sec:exp_further_analysis}

Beyond the latent-design ablations above, we further analyze categorical design choices of \method{} and how its gain scales with data, under the same controlled $\pi_{0.5}$/RoboTwin~2.0 protocol as Section~\ref{sec:exp_ablation}.


\noindent\textbf{Visual-dynamics supervision.}
The visual-dynamics objective $\mathcal{L}_{\mathrm{dyna}}$ is most effective as a dense L1 regression onto a frozen Wan~VAE feature (Fig.~\ref{fig:further_categorical}(a,b)). In panel~(a), Wan~VAE yields the strongest target among the four frozen video encoders, with CogVideoX~\citep{yang2024cogvideox}, SVD~\citep{blattmann2023svd}, and the self-supervised V-JEPA target~\citep{assran2025vjepa2} all narrowing the +\method{} gain. In panel~(b), L1 supervises the action latent with per-feature regression on the entire future video-VAE map, a denser signal than the contrastive InfoNCE objective~\citep{oord2018cpc} that only enforces global similarity; replacing L1 with InfoNCE or stacking the two both degrade performance.

\noindent\textbf{Policy prediction target.}
The frozen \method{} latent alone is the most effective policy target (Fig.~\ref{fig:further_categorical}(c)). A raw-only head reproduces the $\pi_{0.5}$ baseline~\citep{intelligence2025pi05}, while a joint head sits in between regardless of which branch decodes at inference. Splitting head capacity between two targets dilutes the supervision on the dynamics-grounded latent, itself a compact target the flow-matching expert fits on its own.

\begin{figure}[t]
  \centering
  \includegraphics[width=\linewidth]{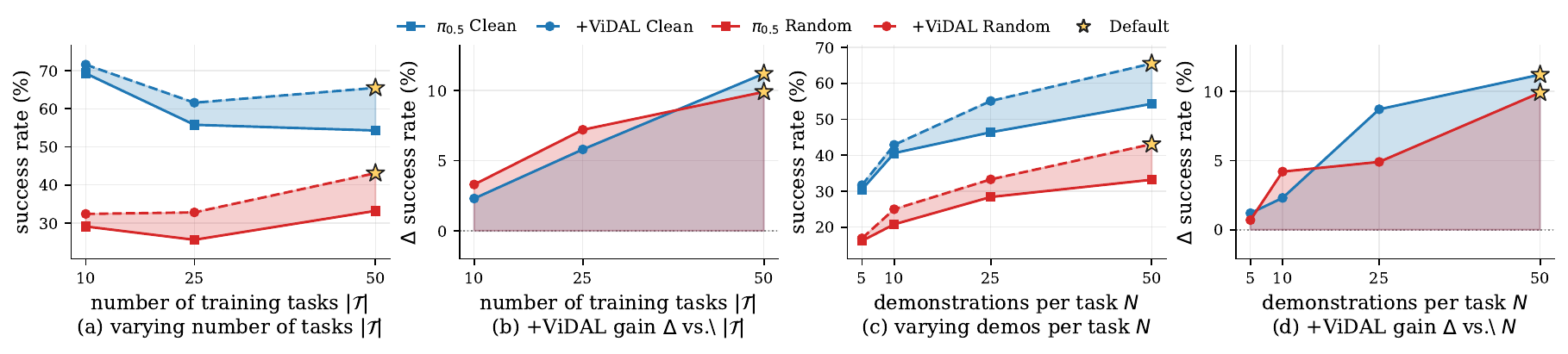}
  \caption{Scaling analysis of \method{} on RoboTwin~2.0 with $\pi_{0.5}$. (a)~success rate against the number of training tasks $|\mathcal{T}|$ (eval subset varies with $|\mathcal{T}|$); (b)~the corresponding success rate improvement $\Delta$ (+\method{}-$\pi_{0.5}$), which is comparable across $|\mathcal{T}|$; (c)~success rate against demonstrations per task $N$ on a fixed 50-task eval; (d)~the corresponding success rate improvement $\Delta$ against $N$.}
  \label{fig:further_scaling}
\end{figure}

\noindent\textbf{Scaling with tasks and demonstrations.}
\method{} improves $\pi_{0.5}$ at every $|\mathcal{T}|$ and $N$ tested, and the +\method{} gain widens as data scales up (Fig.~\ref{fig:further_scaling}). Since the eval subset varies with $|\mathcal{T}|$, the comparable trend is read from $\Delta$ in panel~(b), where the gain widens monotonically; panel~(d) shows the same widening of $\Delta$ along $N$. Greater task diversity exposes the action latent to a broader distribution of visual dynamics, while more demonstrations sharpen each anchoring target, so \method{} is most beneficial in the large-data regime where VLAs are typically trained.

\subsection{Real-Robot Experiments}
\label{sec:exp_real_world}

\begin{figure}[t]
  \centering
  \includegraphics[width=\linewidth]{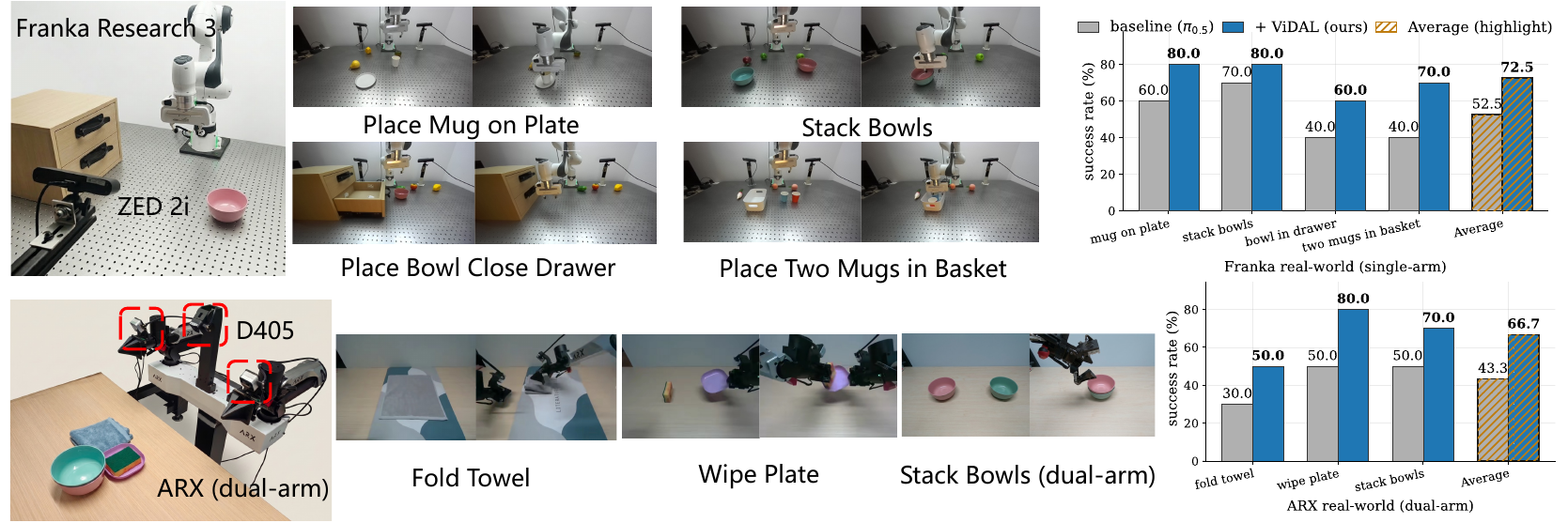}
  \caption{Real-world deployment of \method{} on two robot platforms. \textit{Top:} a single-arm Franka Research~3 with a ZED~2i RGB-D camera on four tabletop tasks. \textit{Bottom:} a dual-arm ARX with three RealSense D405 cameras on three bimanual tasks. Each panel shows the hardware setup, representative rollouts, and success rates of $\pi_{0.5}$ with and without \method{} over 10 rollouts per task.}
  \label{fig:real_world}
\end{figure}

\noindent\textbf{Setup.}
We deploy \method{} on two real-robot platforms to test whether its simulation gains transfer to physical hardware (Fig.~\ref{fig:real_world}). The single-arm setup is a Franka Research~3 with a ZED~2i RGB-D camera on four tabletop tasks (\textit{Place Mug on Plate}, \textit{Stack Bowls}, \textit{Place Bowl Close Drawer}, \textit{Place Two Mugs in Basket}), and the dual-arm setup is an ARX with three RealSense D405 cameras on three bimanual tasks (\textit{Fold Towel}, \textit{Wipe Plate}, \textit{Stack Bowls (dual-arm)}). On each platform, the Action VAE is trained on that platform's own demonstrations and reused as a frozen plug-in action representation, while the underlying $\pi_{0.5}$ policy and its training recipe are kept identical to the simulation runs; the only difference between the two compared variants is whether $\pi_{0.5}$ predicts raw actions or \method{} latents. Each task is evaluated over 10 rollouts and reported as success rate.

\noindent\textbf{Results.}
\method{} consistently improves $\pi_{0.5}$ on every task across both platforms, raising the average success rate by $+20.0\%$ on the single-arm Franka and $+23.4\%$ on the dual-arm ARX (Fig.~\ref{fig:real_world}). The largest single-arm gains appear on tasks that require precise sequencing or insertion under geometric constraints (\textit{Place Bowl Close Drawer}, \textit{Place Two Mugs in Basket}), where the baseline often misjudges approach timing or contact pose; the visually anchored action latent better aligns the predicted action with the resulting scene change, narrowing this failure mode. On the dual-arm side, both deformable manipulation (\textit{Fold Towel}) and rigid bimanual coordination (\textit{Wipe Plate}, \textit{Stack Bowls (dual-arm)}) benefit, and the average gain even exceeds that of the single-arm setup, suggesting that the dynamics-grounded latent is especially useful when contact-rich, higher-DoF behavior must be coordinated across two arms. Together, these results confirm that visual-dynamics grounding transfers cleanly from simulation to real robots, without any retraining or recipe changes to the underlying VLA policy.

\subsection{Qualitative Visualization and Analysis}
\label{sec:exp_qualitative}

We examine the learned action latent space qualitatively, asking whether \method{} yields a richer latent geometry than raw actions and whether it retains enough visual-dynamics information to reconstruct future video frames.

\begin{figure}[t]
  \centering
  \includegraphics[width=\linewidth]{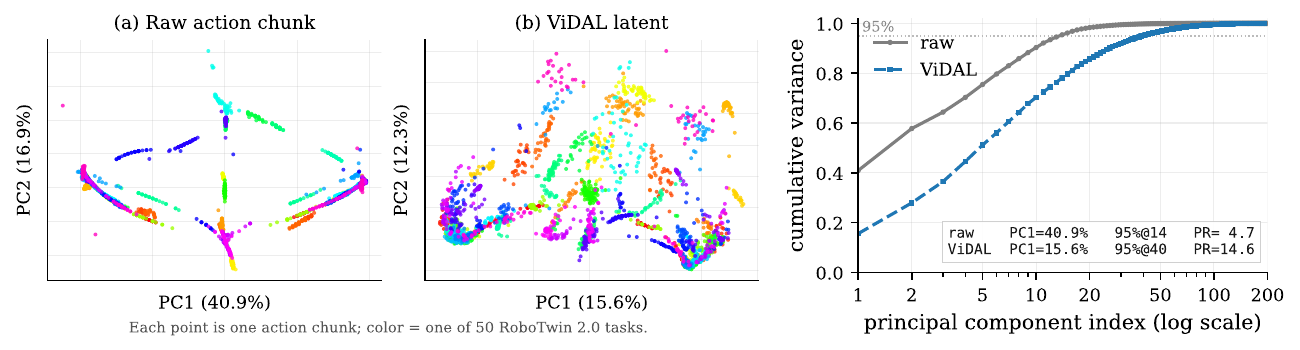}
  \caption{Action latent space geometry on $50$ RoboTwin~2.0 tasks. \textit{Left/middle:} 2D PCA projections of raw action chunks and \method{} latents, colored by task. \textit{Right:} cumulative explained variance vs.\ principal-component index.}
  \label{fig:pca}
\end{figure}

\noindent\textbf{Action latent space geometry.}
\method{} produces a less collapsed action latent space than raw actions (Fig.~\ref{fig:pca}). The top principal component absorbs $40.9\%$ of total variance for raw action chunks but only $15.6\%$ for \method{}, and reaching $95\%$ cumulative variance needs $14$ versus $40$ components. The participation ratio~\citep{gao2015simplicity}, the effective number of significant latent directions, rises from $4.7$ to $14.6$ ($3.1\times$). The 2D PCA projections confirm this: raw chunks collapse onto a few low-dimensional curves, whereas \method{} latents spread across more directions and separate tasks more uniformly. Such a well-regularized latent is the kind of representation that flow-matching action experts learn effectively from~\citep{rombach2022ldm,karras2024edm2}.

\begin{figure}[t]
  \centering
  \includegraphics[width=\linewidth]{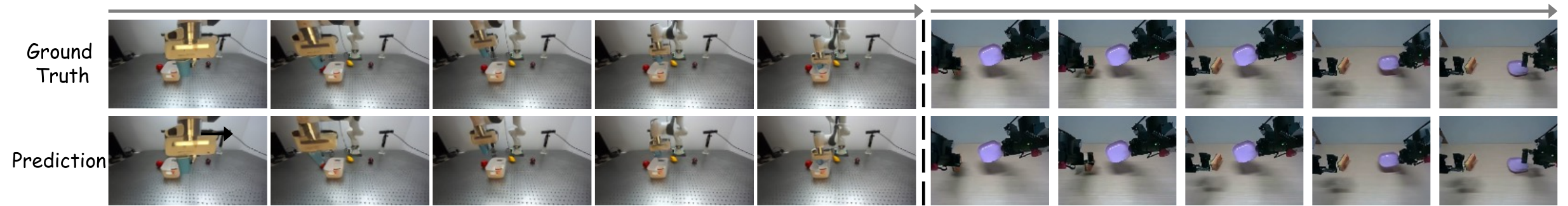}
  \caption{Future video prediction from the \method{} action latent. For each episode (\textit{left:} real-world Franka; \textit{right:} real-world ARX), top: ground-truth future frames; bottom: frames decoded via the future visual latent predictor and frozen video VAE decoder.}
  \label{fig:pred}
\end{figure}

\noindent\textbf{Future video prediction.}
The same action latent the VLA policy predicts also drives a recognizable future video rollout (Fig.~\ref{fig:pred}). Feeding $(\mathbf{z}_0,\mathbf{z}_a)$ through the future visual latent predictor $P_\theta$ and decoding $\hat{\mathbf{z}}_f$ with the frozen video VAE decoder yields frames that track the ground truth in object location, gripper pose, and contact events. The action latent therefore encodes what the scene should look like next, not only what the robot should do. Both the predictor and video decoder are used only for analysis and are not required during policy inference.

\section{Conclusion}

We propose \method{}, a method that grounds the action latent space in future visual dynamics by jointly training an Action VAE on action reconstruction and video latent prediction. The frozen Action VAE then serves as a plug-in action representation for both regression and flow-matching VLA policies without modifying the VLM backbone, and consistently improves performance across two simulation benchmarks (LIBERO, RoboTwin~2.0) and two real-robot platforms (single-arm Franka, dual-arm ARX), and further analysis confirms that the advantage widens as tasks and demonstrations scale up. Since \method{} decouples action-representation learning from VLA policy training, extending it to diverse action spaces and embodiments requires only retraining the lightweight Action VAE; leveraging large-scale egocentric human videos as an additional source of visual-dynamics supervision is a natural next step toward a universal dynamics-grounded action latent space.

{
  \small
  \bibliographystyle{unsrtnat}
  \bibliography{references}

\begin{thebibliography}{43}
\providecommand{\natexlab}[1]{#1}
\providecommand{\url}[1]{\texttt{#1}}
\expandafter\ifx\csname urlstyle\endcsname\relax
  \providecommand{\doi}[1]{doi: #1}\else
  \providecommand{\doi}{doi: \begingroup \urlstyle{rm}\Url}\fi

\bibitem[Brohan et~al.(2022)Brohan, Brown, Carbajal, Chebotar, Dabis, Finn,
  Gopalakrishnan, Hausman, Herzog, Hsu, et~al.]{brohan2022rt1}
Anthony Brohan, Noah Brown, Justice Carbajal, Yevgen Chebotar, Joseph Dabis,
  Chelsea Finn, Keerthana Gopalakrishnan, Karol Hausman, Alex Herzog, Jasmine
  Hsu, et~al.
\newblock Rt-1: Robotics transformer for real-world control at scale.
\newblock \emph{arXiv preprint arXiv:2212.06817}, 2022.

\bibitem[Zitkovich et~al.(2023)Zitkovich, Yu, Xu, Xu, Xiao, Xia, Wu, Wohlhart,
  Welker, Wahid, et~al.]{zitkovich2023rt2}
Brianna Zitkovich, Tianhe Yu, Sichun Xu, Peng Xu, Ted Xiao, Fei Xia, Jialin Wu,
  Paul Wohlhart, Stefan Welker, Ayzaan Wahid, et~al.
\newblock Rt-2: Vision-language-action models transfer web knowledge to robotic
  control.
\newblock In \emph{Conference on Robot Learning}, pages 2165--2183. PMLR, 2023.

\bibitem[Kim et~al.(2024)Kim, Pertsch, Karamcheti, Xiao, Balakrishna, Nair,
  Rafailov, Foster, Lam, Sanketi, et~al.]{kim2024openvla}
Moo~Jin Kim, Karl Pertsch, Siddharth Karamcheti, Ted Xiao, Ashwin Balakrishna,
  Suraj Nair, Rafael Rafailov, Ethan Foster, Grace Lam, Pannag Sanketi, et~al.
\newblock Openvla: An open-source vision-language-action model.
\newblock \emph{arXiv preprint arXiv:2406.09246}, 2024.

\bibitem[Black et~al.(2024)Black, Brown, Driess, Esmail, Equi, Finn, Fusai,
  Groom, Hausman, Ichter, et~al.]{black2024pi0}
Kevin Black, Noah Brown, Danny Driess, Adnan Esmail, Michael Equi, Chelsea
  Finn, Niccolo Fusai, Lachy Groom, Karol Hausman, Brian Ichter, et~al.
\newblock {$\pi_0$}: A vision-language-action flow model for general robot
  control.
\newblock \emph{arXiv preprint arXiv:2410.24164}, 2024.

\bibitem[Intelligence et~al.(2025)Intelligence, Black, Brown, Darpinian,
  Dhabalia, Driess, Esmail, Equi, Finn, Fusai, Galliker, Ghosh, Groom, Hausman,
  Ichter, Jakubczak, Jones, Ke, LeBlanc, Levine, Li-Bell, Mothukuri, Nair,
  Pertsch, Ren, Shi, Smith, Springenberg, Stachowicz, Tanner, Vuong, Walke,
  Walling, Wang, Yu, and Zhilinsky]{intelligence2025pi05}
Physical Intelligence, Kevin Black, Noah Brown, James Darpinian, Karan
  Dhabalia, Danny Driess, Adnan Esmail, Michael Equi, Chelsea Finn, Niccolo
  Fusai, Manuel~Y. Galliker, Dibya Ghosh, Lachy Groom, Karol Hausman, Brian
  Ichter, Szymon Jakubczak, Tim Jones, Liyiming Ke, Devin LeBlanc, Sergey
  Levine, Adrian Li-Bell, Mohith Mothukuri, Suraj Nair, Karl Pertsch, Allen~Z.
  Ren, Lucy~Xiaoyang Shi, Laura Smith, Jost~Tobias Springenberg, Kyle
  Stachowicz, James Tanner, Quan Vuong, Homer Walke, Anna Walling, Haohuan
  Wang, Lili Yu, and Ury Zhilinsky.
\newblock {$\pi_{0.5}$}: A vision-language-action model with open-world
  generalization.
\newblock \emph{arXiv preprint arXiv:2504.16054}, 2025.

\bibitem[Liu et~al.(2024)Liu, Wu, Li, Tan, Chen, Wang, Xu, Su, and
  Zhu]{liu2024rdt}
Songming Liu, Lingxuan Wu, Bangguo Li, Hengkai Tan, Huayu Chen, Zhengyi Wang,
  Ke~Xu, Hang Su, and Jun Zhu.
\newblock Rdt-1b: a diffusion foundation model for bimanual manipulation.
\newblock \emph{arXiv preprint arXiv:2410.07864}, 2024.

\bibitem[Chi et~al.(2025)Chi, Xu, Feng, Cousineau, Du, Burchfiel, Tedrake, and
  Song]{chi2023diffusion}
Cheng Chi, Zhenjia Xu, Siyuan Feng, Eric Cousineau, Yilun Du, Benjamin
  Burchfiel, Russ Tedrake, and Shuran Song.
\newblock Diffusion policy: Visuomotor policy learning via action diffusion.
\newblock \emph{The International Journal of Robotics Research}, 44\penalty0
  (10-11):\penalty0 1684--1704, 2025.

\bibitem[Pertsch et~al.(2025)Pertsch, Stachowicz, Ichter, Driess, Nair, Vuong,
  Mees, Finn, and Levine]{pertsch2025fast}
Karl Pertsch, Kyle Stachowicz, Brian Ichter, Danny Driess, Suraj Nair, Quan
  Vuong, Oier Mees, Chelsea Finn, and Sergey Levine.
\newblock Fast: Efficient action tokenization for vision-language-action
  models.
\newblock \emph{arXiv preprint arXiv:2501.09747}, 2025.

\bibitem[Wang et~al.(2025)Wang, Zhu, Liu, Yang, Fang, and He]{wang2025vqvla}
Yating Wang, Haoyi Zhu, Mingyu Liu, Jiange Yang, Hao-Shu Fang, and Tong He.
\newblock Vq-vla: Improving vision-language-action models via scaling
  vector-quantized action tokenizers.
\newblock pages 11089--11099, 2025.

\bibitem[Zhou et~al.(2025)Zhou, Liao, Huang, Tang, Otto, Jia, Jiang, Hilber,
  Li, Wang, et~al.]{zhou2025beast}
Hongyi Zhou, Weiran Liao, Xi~Huang, Yucheng Tang, Fabian Otto, Xiaogang Jia,
  Xinkai Jiang, Simon Hilber, Ge~Li, Qian Wang, et~al.
\newblock Beast: Efficient tokenization of b-splines encoded action sequences
  for imitation learning.
\newblock \emph{arXiv preprint arXiv:2506.06072}, 2025.

\bibitem[Liu et~al.(2025)Liu, Zhang, Dong, Ye, Yuan, Yu, Yin, Lu, Shi, Yu,
  et~al.]{faster2025}
Yicheng Liu, Shiduo Zhang, Zibin Dong, Baijun Ye, Tianyuan Yuan, Xiaopeng Yu,
  Linqi Yin, Chenhao Lu, Junhao Shi, Luca Jiang-Tao Yu, et~al.
\newblock Faster: Toward efficient autoregressive vision language action
  modeling via neural action tokenization.
\newblock \emph{arXiv preprint arXiv:2512.04952}, 2025.

\bibitem[Tan et~al.(2024)Tan, Liu, Zhang, Song, and Fu]{tan2024rold}
Wenhui Tan, Bei Liu, Junbo Zhang, Ruihua Song, and Jianlong Fu.
\newblock Rold: Robot latent diffusion for multi-task policy modeling.
\newblock pages 340--353, 2024.

\bibitem[Walke et~al.(2023)Walke, Black, Zhao, Vuong, Zheng, Hansen-Estruch,
  He, Myers, Kim, Du, et~al.]{walke2023bridgedata}
Homer~Rich Walke, Kevin Black, Tony~Z Zhao, Quan Vuong, Chongyi Zheng, Philippe
  Hansen-Estruch, Andre~Wang He, Vivek Myers, Moo~Jin Kim, Max Du, et~al.
\newblock Bridgedata v2: A dataset for robot learning at scale.
\newblock In \emph{Conference on Robot Learning}, pages 1723--1736. PMLR, 2023.

\bibitem[O’Neill et~al.(2024)O’Neill, Rehman, Maddukuri, Gupta, Padalkar,
  Lee, Pooley, Gupta, Mandlekar, Jain, et~al.]{openx2023}
Abby O’Neill, Abdul Rehman, Abhiram Maddukuri, Abhishek Gupta, Abhishek
  Padalkar, Abraham Lee, Acorn Pooley, Agrim Gupta, Ajay Mandlekar, Ajinkya
  Jain, et~al.
\newblock Open x-embodiment: Robotic learning datasets and rt-x models: Open
  x-embodiment collaboration 0.
\newblock pages 6892--6903, 2024.

\bibitem[Khazatsky et~al.(2024)Khazatsky, Pertsch, Nair, Balakrishna, Dasari,
  Karamcheti, Nasiriany, Srirama, Chen, Ellis, et~al.]{khazatsky2024droid}
Alexander Khazatsky, Karl Pertsch, Suraj Nair, Ashwin Balakrishna, Sudeep
  Dasari, Siddharth Karamcheti, Soroush Nasiriany, Mohan~Kumar Srirama,
  Lawrence~Yunliang Chen, Kirsty Ellis, et~al.
\newblock Droid: A large-scale in-the-wild robot manipulation dataset.
\newblock \emph{arXiv preprint arXiv:2403.12945}, 2024.

\bibitem[Team et~al.(2024)Team, Ghosh, Walke, Pertsch, Black, Mees, Dasari,
  Hejna, Kreiman, Xu, et~al.]{octo2024}
Octo~Model Team, Dibya Ghosh, Homer Walke, Karl Pertsch, Kevin Black, Oier
  Mees, Sudeep Dasari, Joey Hejna, Tobias Kreiman, Charles Xu, et~al.
\newblock Octo: An open-source generalist robot policy.
\newblock 2024.

\bibitem[Assran et~al.(2025)Assran, Bardes, Fan, Garrido, Howes, Muckley,
  Rizvi, Roberts, Sinha, Zholus, et~al.]{assran2025vjepa2}
Mido Assran, Adrien Bardes, David Fan, Quentin Garrido, Russell Howes, Matthew
  Muckley, Ammar Rizvi, Claire Roberts, Koustuv Sinha, Artem Zholus, et~al.
\newblock V-jepa 2: Self-supervised video models enable understanding,
  prediction and planning.
\newblock \emph{arXiv preprint arXiv:2506.09985}, 2025.

\bibitem[Hu et~al.(2024)Hu, Guo, Wang, Chen, Wang, Zhang, Sreenath, Lu, and
  Chen]{hu2025vpp}
Yucheng Hu, Yanjiang Guo, Pengchao Wang, Xiaoyu Chen, Yen-Jen Wang, Jianke
  Zhang, Koushil Sreenath, Chaochao Lu, and Jianyu Chen.
\newblock Video prediction policy: A generalist robot policy with predictive
  visual representations.
\newblock 2024.

\bibitem[Cen et~al.(2025)Cen, Yu, Yuan, Jiang, Huang, Guo, Li, Song, Luo, Wang,
  et~al.]{cen2025worldvla}
Jun Cen, Chaohui Yu, Hangjie Yuan, Yuming Jiang, Siteng Huang, Jiayan Guo, Xin
  Li, Yibing Song, Hao Luo, Fan Wang, et~al.
\newblock Worldvla: Towards autoregressive action world model.
\newblock \emph{arXiv preprint arXiv:2506.21539}, 2025.

\bibitem[Xu et~al.(2026)Xu, Li, Ye, Chen, Zeng, Chen, Xu, Lin, Li, and
  Pang]{xu2026futurevla}
Xiaoxu Xu, Hao Li, Jinhui Ye, Yilun Chen, Jia Zeng, Xinyi Chen, Linning Xu,
  Dahua Lin, Weixin Li, and Jiangmiao Pang.
\newblock Futurevla: Joint visuomotor prediction for vision-language-action
  model.
\newblock \emph{arXiv preprint arXiv:2603.10712}, 2026.

\bibitem[Shang et~al.(2026)Shang, Zhan, Yan, Wang, Li, An, Wang, Liu, Hou, Fan,
  et~al.]{shang2026dynvla}
Shuyao Shang, Bing Zhan, Yunfei Yan, Yuqi Wang, Yingyan Li, Yasong An, Xiaoman
  Wang, Jierui Liu, Lu~Hou, Lue Fan, et~al.
\newblock Dynvla: Learning world dynamics for action reasoning in autonomous
  driving.
\newblock \emph{arXiv preprint arXiv:2603.11041}, 2026.

\bibitem[Su et~al.(2026)Su, Chen, Shi, Liu, Zhang, Huang, Zhong, Zhu, Liu, and
  Liu]{su2026worldguidance}
Yue Su, Sijin Chen, Haixin Shi, Mingyu Liu, Zhengshen Zhang, Ningyuan Huang,
  Weiheng Zhong, Zhengbang Zhu, Yuxiao Liu, and Xihui Liu.
\newblock World guidance: World modeling in condition space for action
  generation.
\newblock \emph{arXiv preprint arXiv:2602.22010}, 2026.

\bibitem[Kim et~al.(2025)Kim, Finn, and Liang]{kim2025openvlaoft}
Moo~Jin Kim, Chelsea Finn, and Percy Liang.
\newblock Fine-tuning vision-language-action models: Optimizing speed and
  success.
\newblock \emph{arXiv preprint arXiv:2502.19645}, 2025.

\bibitem[Liu et~al.(2023)Liu, Zhu, Gao, Feng, Liu, Zhu, and
  Stone]{liu2024libero}
Bo~Liu, Yifeng Zhu, Chongkai Gao, Yihao Feng, Qiang Liu, Yuke Zhu, and Peter
  Stone.
\newblock Libero: Benchmarking knowledge transfer for lifelong robot learning.
\newblock \emph{Advances in Neural Information Processing Systems},
  36:\penalty0 44776--44791, 2023.

\bibitem[Chen et~al.(2025{\natexlab{a}})Chen, Chen, Chen, Cai, Liu, Li, Liang,
  Lin, Ge, Gu, et~al.]{mu2024robotwin}
Tianxing Chen, Zanxin Chen, Baijun Chen, Zijian Cai, Yibin Liu, Zixuan Li,
  Qiwei Liang, Xianliang Lin, Yiheng Ge, Zhenyu Gu, et~al.
\newblock Robotwin 2.0: A scalable data generator and benchmark with strong
  domain randomization for robust bimanual robotic manipulation.
\newblock 2025{\natexlab{a}}.

\bibitem[Ye et~al.(2026{\natexlab{a}})Ye, Gao, Yang, Zheng, Wang, Chen, Chen,
  Chen, Liu, and Jia]{ye2026starvlaalpha}
Jinhui Ye, Ning Gao, Senqiao Yang, Jinliang Zheng, Zixuan Wang, Yuxin Chen,
  Pengguang Chen, Yilun Chen, Shu Liu, and Jiaya Jia.
\newblock {StarVLA-$\alpha$}: Reducing complexity in vision-language-action
  systems.
\newblock \emph{arXiv preprint arXiv:2604.11757}, 2026{\natexlab{a}}.

\bibitem[Driess et~al.(2023)Driess, Xia, Sajjadi, Lynch, Chowdhery, Ichter,
  Wahid, Tompson, Vuong, Yu, et~al.]{driess2023palme}
Danny Driess, Fei Xia, Mehdi~SM Sajjadi, Corey Lynch, Aakanksha Chowdhery,
  Brian Ichter, Ayzaan Wahid, Jonathan Tompson, Quan Vuong, Tianhe Yu, et~al.
\newblock Palm-e: An embodied multimodal language model.
\newblock 2023.

\bibitem[Li et~al.(2023)Li, Liu, Zhang, Yu, Xu, Wu, Cheang, Jing, Zhang, Liu,
  et~al.]{li2024roboflamingo}
Xinghang Li, Minghuan Liu, Hanbo Zhang, Cunjun Yu, Jie Xu, Hongtao Wu, Chilam
  Cheang, Ya~Jing, Weinan Zhang, Huaping Liu, et~al.
\newblock Vision-language foundation models as effective robot imitators.
\newblock 2023.

\bibitem[Zhao et~al.(2023)Zhao, Kumar, Levine, and Finn]{zhao2023act}
Tony~Z Zhao, Vikash Kumar, Sergey Levine, and Chelsea Finn.
\newblock Learning fine-grained bimanual manipulation with low-cost hardware.
\newblock \emph{arXiv preprint arXiv:2304.13705}, 2023.

\bibitem[Sun et~al.(2026)Sun, Zhang, Qi, Ren, Liu, Zhu, Sun, Jin, and
  Chen]{sun2026vlajepa}
Jingwen Sun, Wenyao Zhang, Zekun Qi, Shaojie Ren, Zezhi Liu, Hanxin Zhu,
  Guangzhong Sun, Xin Jin, and Zhibo Chen.
\newblock Vla-jepa: Enhancing vision-language-action model with latent world
  model.
\newblock \emph{arXiv preprint arXiv:2602.10098}, 2026.

\bibitem[Jia et~al.(2025)Jia, Liu, Liu, Zhou, Yu, Yan, Chi, Guo, Shi, and
  Zhang]{jia2025video2act}
Yueru Jia, Jiaming Liu, Shengbang Liu, Rui Zhou, Wanhe Yu, Yuyang Yan, Xiaowei
  Chi, Yandong Guo, Boxin Shi, and Shanghang Zhang.
\newblock Video2act: A dual-system video diffusion policy with robotic
  spatio-motional modeling.
\newblock \emph{arXiv preprint arXiv:2512.03044}, 2025.

\bibitem[Ye et~al.(2026{\natexlab{b}})Ye, Ge, Zheng, Gao, Yu, Kurian, Indupuru,
  Tan, Zhu, Xiang, et~al.]{dreamzero2026}
Seonghyeon Ye, Yunhao Ge, Kaiyuan Zheng, Shenyuan Gao, Sihyun Yu, George
  Kurian, Suneel Indupuru, You~Liang Tan, Chuning Zhu, Jiannan Xiang, et~al.
\newblock World action models are zero-shot policies.
\newblock \emph{arXiv preprint arXiv:2602.15922}, 2026{\natexlab{b}}.

\bibitem[Bjorck et~al.(2025)Bjorck, Casta{\~n}eda, Cherniadev, Da, Ding, Fan,
  Fang, Fox, Hu, Huang, et~al.]{bjorck2025gr00t}
Johan Bjorck, Fernando Casta{\~n}eda, Nikita Cherniadev, Xingye Da, Runyu Ding,
  Linxi Fan, Yu~Fang, Dieter Fox, Fengyuan Hu, Spencer Huang, et~al.
\newblock Gr00t n1: An open foundation model for generalist humanoid robots.
\newblock \emph{arXiv preprint arXiv:2503.14734}, 2025.

\bibitem[Zhang et~al.(2025)Zhang, Liu, Qi, Wang, Yu, Zhang, Dong, He, Lu, Wang,
  et~al.]{zhang2025dreamvla}
Wenyao Zhang, Hongsi Liu, Zekun Qi, Yunnan Wang, Xinqiang Yu, Jiazhao Zhang,
  Runpei Dong, Jiawei He, Fan Lu, He~Wang, et~al.
\newblock Dreamvla: a vision-language-action model dreamed with comprehensive
  world knowledge.
\newblock \emph{arXiv preprint arXiv:2507.04447}, 2025.

\bibitem[Bu et~al.(2025)Bu, Yang, Cai, Gao, Ren, Yao, Luo, and
  Li]{bu2025univla}
Qingwen Bu, Yanting Yang, Jisong Cai, Shenyuan Gao, Guanghui Ren, Maoqing Yao,
  Ping Luo, and Hongyang Li.
\newblock Univla: Learning to act anywhere with task-centric latent actions.
\newblock \emph{arXiv preprint arXiv:2505.06111}, 2025.

\bibitem[Chen et~al.(2025{\natexlab{b}})Chen, Wei, Zhang, Zhang, Wang, Guo,
  Yang, Wang, Xiao, Zhao, et~al.]{chen2025villa}
Xiaoyu Chen, Hangxing Wei, Pushi Zhang, Chuheng Zhang, Kaixin Wang, Yanjiang
  Guo, Rushuai Yang, Yucen Wang, Xinquan Xiao, Li~Zhao, et~al.
\newblock Villa-x: enhancing latent action modeling in vision-language-action
  models.
\newblock \emph{arXiv preprint arXiv:2507.23682}, 2025{\natexlab{b}}.

\bibitem[Wan et~al.(2025)Wan, Wang, Ai, Wen, Mao, Xie, Chen, Yu, Zhao, Yang,
  Zeng, Wang, Zhang, Zhou, Wang, Chen, Zhu, Zhao, Yan, Huang, Feng, Zhang, Li,
  Wu, Chu, Feng, Zhang, Sun, Fang, Wang, Gui, Weng, Shen, Lin, Wang, Wang,
  Zhou, Wang, Shen, Yu, Shi, Huang, Xu, Kou, Lv, Li, Liu, Wang, Zhang, Huang,
  Li, Wu, Liu, Pan, Zheng, Hong, Shi, Feng, Jiang, Han, Wu, and
  Liu]{wan2025wan}
Team Wan, Ang Wang, Baole Ai, Bin Wen, Chaojie Mao, Chen-Wei Xie, Di~Chen,
  Feiwu Yu, Haiming Zhao, Jianxiao Yang, Jianyuan Zeng, Jiayu Wang, Jingfeng
  Zhang, Jingren Zhou, Jinkai Wang, Jixuan Chen, Kai Zhu, Kang Zhao, Keyu Yan,
  Lianghua Huang, Mengyang Feng, Ningyi Zhang, Pandeng Li, Pingyu Wu, Ruihang
  Chu, Ruili Feng, Shiwei Zhang, Siyang Sun, Tao Fang, Tianxing Wang, Tianyi
  Gui, Tingyu Weng, Tong Shen, Wei Lin, Wei Wang, Wei Wang, Wenmeng Zhou, Wente
  Wang, Wenting Shen, Wenyuan Yu, Xianzhong Shi, Xiaoming Huang, Xin Xu, Yan
  Kou, Yangyu Lv, Yifei Li, Yijing Liu, Yiming Wang, Yingya Zhang, Yitong
  Huang, Yong Li, You Wu, Yu~Liu, Yulin Pan, Yun Zheng, Yuntao Hong, Yupeng
  Shi, Yutong Feng, Zeyinzi Jiang, Zhen Han, Zhi-Fan Wu, and Ziyu Liu.
\newblock Wan: Open and advanced large-scale video generative models.
\newblock \emph{arXiv preprint arXiv:2503.20314}, 2025.

\bibitem[Yang et~al.(2024)Yang, Teng, Zheng, Ding, Huang, Xu, Yang, Hong,
  Zhang, Feng, et~al.]{yang2024cogvideox}
Zhuoyi Yang, Jiayan Teng, Wendi Zheng, Ming Ding, Shiyu Huang, Jiazheng Xu,
  Yuanming Yang, Wenyi Hong, Xiaohan Zhang, Guanyu Feng, et~al.
\newblock Cogvideox: Text-to-video diffusion models with an expert transformer.
\newblock \emph{arXiv preprint arXiv:2408.06072}, 2024.

\bibitem[Blattmann et~al.(2023)Blattmann, Dockhorn, Kulal, Mendelevitch,
  Kilian, Lorenz, Levi, English, Voleti, Letts, et~al.]{blattmann2023svd}
Andreas Blattmann, Tim Dockhorn, Sumith Kulal, Daniel Mendelevitch, Maciej
  Kilian, Dominik Lorenz, Yam Levi, Zion English, Vikram Voleti, Adam Letts,
  et~al.
\newblock Stable video diffusion: Scaling latent video diffusion models to
  large datasets.
\newblock \emph{arXiv preprint arXiv:2311.15127}, 2023.

\bibitem[Oord et~al.(2018)Oord, Li, and Vinyals]{oord2018cpc}
Aaron van~den Oord, Yazhe Li, and Oriol Vinyals.
\newblock Representation learning with contrastive predictive coding.
\newblock \emph{arXiv preprint arXiv:1807.03748}, 2018.

\bibitem[Gao and Ganguli(2015)]{gao2015simplicity}
Peiran Gao and Surya Ganguli.
\newblock On simplicity and complexity in the brave new world of large-scale
  neuroscience.
\newblock \emph{Current opinion in neurobiology}, 32:\penalty0 148--155, 2015.

\bibitem[Rombach et~al.(2022)Rombach, Blattmann, Lorenz, Esser, and
  Ommer]{rombach2022ldm}
Robin Rombach, Andreas Blattmann, Dominik Lorenz, Patrick Esser, and Bj{\"o}rn
  Ommer.
\newblock High-resolution image synthesis with latent diffusion models.
\newblock In \emph{Proceedings of the IEEE/CVF conference on computer vision
  and pattern recognition}, pages 10684--10695, 2022.

\bibitem[Karras et~al.(2024)Karras, Aittala, Lehtinen, Hellsten, Aila, and
  Laine]{karras2024edm2}
Tero Karras, Miika Aittala, Jaakko Lehtinen, Janne Hellsten, Timo Aila, and
  Samuli Laine.
\newblock Analyzing and improving the training dynamics of diffusion models.
\newblock In \emph{Proceedings of the IEEE/CVF conference on computer vision
  and pattern recognition}, pages 24174--24184, 2024.

\end{thebibliography}
}

\newpage
\appendix
\section{Method details}
\label{sec:appendix_method_details}

\noindent\textbf{Action VAE architecture.}
The Action VAE maps a normalized action chunk $\mathbf{a}_{t:t+H-1}\in\mathbb{R}^{H\times d_a}$ to a short sequence of continuous action-latent tokens. Its encoder is a temporal convolutional network with residual blocks and strided temporal downsampling, so the latent sequence preserves the temporal order of the action chunk while reducing its length. The encoder predicts a Gaussian posterior with mean and variance parameters, from which the action latent is sampled during representation learning. For downstream VLA supervision, we use the posterior mean as a deterministic action-latent target. The decoder mirrors the encoder with temporal upsampling and reconstructs the original action chunk from the latent sequence. This design keeps the action representation continuous, temporally structured, and directly decodable into executable robot actions.

\noindent\textbf{Frozen video-latent target.}
For each action chunk, we use the aligned observation clip $\mathbf{o}_{t:t+H}$, which contains one more frame than the action horizon. A frozen video VAE encodes this clip into spatiotemporal latent tokens. For a clip with $H{+}1=4k{+}1$ frames, image size $h{\times}w$, and video-latent channel dimension $C_v$, the flattened video latent has shape
\begin{equation}
  \mathbf{z}_v \in \mathbb{R}^{T_v N_s \times C_v},
  \qquad
  T_v = k+1,\quad
  N_s = (h/16)(w/16).
\end{equation}
We split this latent into the first-frame spatial tokens $\mathbf{z}_0\in\mathbb{R}^{N_s\times C_v}$ and the future target $\mathbf{z}_f\in\mathbb{R}^{(T_v-1)N_s\times C_v}$. The first-frame tokens provide the current visual context, while the future tokens provide the visual-dynamics target induced by the action chunk.

\noindent\textbf{Future-latent predictor.}
The dynamics predictor learns to infer the future visual latent from the current visual context and the action latent. It represents the current and future visual positions as a shared token sequence, conditions the future queries on the action-latent tokens through cross-attention, and predicts the future visual-latent tokens jointly:
\begin{equation}
  \hat{\mathbf{z}}_f=P_\theta(\mathbf{z}_0,\mathbf{z}_a).
\end{equation}
Conditioning on $\mathbf{z}_0$ disambiguates actions whose visual effects depend on the current scene, while conditioning on $\mathbf{z}_a$ encourages the action latent to encode the scene transition caused by the action chunk. The predictor is used only during representation learning and is not part of the deployed VLA policy.

\noindent\textbf{Training objective.}
The Action VAE is trained with four terms:
\begin{equation}
  \mathcal{L}_{\text{\method{}}}
  = \mathcal{L}_{\mathrm{rec}}
  + \lambda_{\mathrm{freq}}\mathcal{L}_{\mathrm{freq}}
  + \lambda_{\mathrm{KL}}\mathcal{L}_{\mathrm{KL}}
  + \lambda_{\mathrm{dyna}}\mathcal{L}_{\mathrm{dyna}}.
\end{equation}
The reconstruction term is an L1 loss on normalized actions, $\mathcal{L}_{\mathrm{rec}}=\|\hat{\mathbf{a}}-\mathbf{a}\|_1$. The frequency-domain term applies a discrete cosine transform along the temporal axis and computes an L1 distance between reconstructed and target frequency coefficients. The KL term is the standard Gaussian-posterior regularizer. The visual-dynamics term is an L1 loss on frozen video-latent tokens,
\begin{equation}
  \mathcal{L}_{\mathrm{dyna}}
  = \|P_\theta(\mathbf{z}_0,\mathbf{z}_a)-\mathrm{sg}(\mathbf{z}_f)\|_1,
\end{equation}
where $\mathrm{sg}(\cdot)$ denotes that the video-latent target is held fixed. The weights and schedules are benchmark-specific and are listed in Section~\ref{sec:appendix_implementation_details}.

\noindent\textbf{Downstream VLA interface.}
After representation learning, the video VAE and future-latent predictor are discarded. The frozen Action VAE defines the interface between a VLA policy and the robot: its encoder converts each action chunk into a deterministic latent target, and its decoder converts predicted latents back into executable action chunks. We normalize the action latents before policy training and invert the normalization before decoding at inference. Thus, downstream VLA training changes only the action target space; the VLM backbone and the policy action expert remain otherwise unchanged.

\section{Implementation details}
\label{sec:appendix_implementation_details}

\noindent\textbf{Action VAE training.}
We train a separate Action VAE for each benchmark or robot platform, using the same architecture and objective described in Section~\ref{sec:appendix_method_details}. Table~\ref{tab:appendix_action_vae_config} summarizes the main configuration choices. Single-arm settings use the LIBERO configuration, and the Franka real-robot setting follows the same configuration. Dual-arm settings use the RoboTwin~2.0 configuration, and the ARX real-robot setting follows the same configuration. Across settings, we use a frozen Wan video VAE as the visual-dynamics target, $\lambda_{\mathrm{dyna}}{=}1$, $\lambda_{\mathrm{freq}}{=}1$, and $\lambda_{\mathrm{KL}}{=}10^{-4}$.

\begin{table}[t]
  \caption{Key Action VAE training configurations. We report only the settings that define the action-latent interface and visual-dynamics target.}
  \label{tab:appendix_action_vae_config}
  \centering
  \resizebox{0.78\linewidth}{!}{
  \begin{tabular}{lcc}
    \toprule
    Setting & LIBERO / Franka & RoboTwin~2.0 / ARX \\
    \midrule
    Action dimension $d_a$ & 7 & 14 \\
    Action chunk length $H$ & 32 & 48 \\
    Action latent dimension $D$ & 8 & 16 \\
    Observation clip length & 33 frames & 49 frames \\
    Image resolution & $128{\times}128$ & $96{\times}128$ \\
    Video latent tokens $T_v{\times}N_s$ & $9{\times}64$ & $13{\times}48$ \\
    Training steps & 50k & 50k \\
    \bottomrule
  \end{tabular}}
\end{table}

\noindent\textbf{Downstream VLA integration.}
For all downstream policies, \method{} only changes the action representation: the raw action chunk target is replaced by the normalized Action VAE latent, and the predicted latent is decoded by the frozen Action VAE decoder before execution. The original VLA training and inference recipes are otherwise kept unchanged for OpenVLA-OFT and StarVLA-OFT. For $\pi_{0.5}$, we keep the original flow-matching objective and inference procedure, but set the action expert learning rate to $10{\times}$ the VLM learning rate. This adjustment helps the pretrained action expert adapt from the raw-action distribution used in $\pi_{0.5}$ pretraining to the \method{} latent distribution. At inference, neither the video VAE nor the future-latent predictor is used; the only additional deployed module is the Action VAE decoder, whose latency overhead is negligible compared with the VLA forward pass.

\noindent\textbf{Training data and evaluation.}
On LIBERO, we use the official training demonstrations and official evaluation protocol. On RoboTwin~2.0, we train on the Clean demonstrations from 50 tasks, with 50 episodes per task and 2{,}500 episodes in total; evaluation follows the official Clean and Random settings, with 100 rollouts per task in each setting. For real-robot experiments, the single-arm Franka setting contains four tasks with 50 demonstrations per task, and the dual-arm ARX setting contains three tasks with 100 demonstrations per task. Each real-robot task is evaluated with 10 rollouts, and a rollout is counted as successful if the robot completes the language instruction for that task.

\section{Full RoboTwin 2.0 results}
\label{sec:appendix_robotwin_full_results}

Table~\ref{tab:appendix_robotwin_full} reports the full per-task RoboTwin~2.0 results corresponding to Table~\ref{tab:robotwin_main}. All methods are evaluated under both Clean and Random settings, with 100 rollouts per task in each setting. The representative rows shown in the main paper are a subset of this full table, and the averages match Table~\ref{tab:robotwin_main} up to rounding.

\begin{table}[p]
  \caption{Full RoboTwin~2.0 success rates (\%) over all 50 tasks. Gray columns indicate the corresponding +\method{} variants.}
  \label{tab:appendix_robotwin_full}
  \centering
  \scriptsize
  \setlength{\tabcolsep}{2.5pt}
  \resizebox{\textwidth}{!}{
  \begin{tabular}{lrr|rr|rr|rr|rr}
    \toprule
    & \multicolumn{2}{c|}{$\pi_0$} &
    \multicolumn{2}{c|}{StarVLA-OFT} &
    \multicolumn{2}{c|}{\cellcolor{gray!10}\shortstack{StarVLA-OFT\\+\method{}}} &
    \multicolumn{2}{c|}{$\pi_{0.5}$} &
    \multicolumn{2}{c}{\cellcolor{gray!10}$\pi_{0.5}$+\method{}} \\
    \cmidrule(lr){2-3} \cmidrule(lr){4-5} \cmidrule(lr){6-7} \cmidrule(lr){8-9} \cmidrule(lr){10-11}
    Task & Clean & Rand. & Clean & Rand. & Clean & Rand. & Clean & Rand. & Clean & Rand. \\
    \midrule
    Adjust Bottle & 90.0 & 56.0 & 96.0 & 0.0 & \cellcolor{gray!10}100.0 & \cellcolor{gray!10}5.0 & 84.0 & 73.0 & \cellcolor{gray!10}99.0 & \cellcolor{gray!10}86.0 \\
    Beat Block Hammer & 43.0 & 21.0 & 58.0 & 1.0 & \cellcolor{gray!10}68.0 & \cellcolor{gray!10}2.0 & 68.0 & 19.0 & \cellcolor{gray!10}86.0 & \cellcolor{gray!10}28.0 \\
    Blocks Ranking RGB & 19.0 & 5.0 & 45.0 & 0.0 & \cellcolor{gray!10}9.0 & \cellcolor{gray!10}0.0 & 48.0 & 22.0 & \cellcolor{gray!10}71.0 & \cellcolor{gray!10}41.0 \\
    Blocks Ranking Size & 7.0 & 1.0 & 27.0 & 0.0 & \cellcolor{gray!10}3.0 & \cellcolor{gray!10}0.0 & 26.0 & 7.0 & \cellcolor{gray!10}37.0 & \cellcolor{gray!10}11.0 \\
    Click Alarmclock & 63.0 & 11.0 & 91.0 & 14.0 & \cellcolor{gray!10}81.0 & \cellcolor{gray!10}6.0 & 56.0 & 42.0 & \cellcolor{gray!10}79.0 & \cellcolor{gray!10}79.0 \\
    Click Bell & 44.0 & 3.0 & 94.0 & 4.0 & \cellcolor{gray!10}98.0 & \cellcolor{gray!10}16.0 & 64.0 & 77.0 & \cellcolor{gray!10}53.0 & \cellcolor{gray!10}51.0 \\
    Dump Bin Bigbin & 83.0 & 24.0 & 68.0 & 1.0 & \cellcolor{gray!10}87.0 & \cellcolor{gray!10}7.0 & 86.0 & 54.0 & \cellcolor{gray!10}92.0 & \cellcolor{gray!10}69.0 \\
    Grab Roller & 96.0 & 80.0 & 93.0 & 14.0 & \cellcolor{gray!10}92.0 & \cellcolor{gray!10}20.0 & 96.0 & 87.0 & \cellcolor{gray!10}100.0 & \cellcolor{gray!10}89.0 \\
    Handover Block & 45.0 & 8.0 & 0.0 & 0.0 & \cellcolor{gray!10}27.0 & \cellcolor{gray!10}0.0 & 32.0 & 1.0 & \cellcolor{gray!10}49.0 & \cellcolor{gray!10}13.0 \\
    Handover Mic & 98.0 & 13.0 & 39.0 & 0.0 & \cellcolor{gray!10}90.0 & \cellcolor{gray!10}1.0 & 75.0 & 26.0 & \cellcolor{gray!10}100.0 & \cellcolor{gray!10}54.0 \\
    Hanging Mug & 11.0 & 3.0 & 15.0 & 0.0 & \cellcolor{gray!10}4.0 & \cellcolor{gray!10}0.0 & 9.0 & 8.0 & \cellcolor{gray!10}16.0 & \cellcolor{gray!10}11.0 \\
    Lift Pot & 84.0 & 36.0 & 0.0 & 0.0 & \cellcolor{gray!10}56.0 & \cellcolor{gray!10}0.0 & 58.0 & 2.0 & \cellcolor{gray!10}89.0 & \cellcolor{gray!10}4.0 \\
    Move Can Pot & 58.0 & 21.0 & 50.0 & 0.0 & \cellcolor{gray!10}71.0 & \cellcolor{gray!10}2.0 & 30.0 & 11.0 & \cellcolor{gray!10}62.0 & \cellcolor{gray!10}28.0 \\
    Move Pillbottle Pad & 21.0 & 1.0 & 54.0 & 0.0 & \cellcolor{gray!10}72.0 & \cellcolor{gray!10}0.0 & 35.0 & 17.0 & \cellcolor{gray!10}47.0 & \cellcolor{gray!10}23.0 \\
    Move Playingcard Away & 53.0 & 22.0 & 69.0 & 1.0 & \cellcolor{gray!10}93.0 & \cellcolor{gray!10}5.0 & 64.0 & 35.0 & \cellcolor{gray!10}73.0 & \cellcolor{gray!10}68.0 \\
    Move Stapler Pad & 0.0 & 2.0 & 12.0 & 0.0 & \cellcolor{gray!10}5.0 & \cellcolor{gray!10}0.0 & 23.0 & 12.0 & \cellcolor{gray!10}12.0 & \cellcolor{gray!10}13.0 \\
    Open Laptop & 85.0 & 46.0 & 31.0 & 0.0 & \cellcolor{gray!10}79.0 & \cellcolor{gray!10}6.0 & 75.0 & 48.0 & \cellcolor{gray!10}91.0 & \cellcolor{gray!10}77.0 \\
    Open Microwave & 80.0 & 50.0 & -- & 2.0 & \cellcolor{gray!10}60.0 & \cellcolor{gray!10}15.0 & 63.0 & 31.0 & \cellcolor{gray!10}75.0 & \cellcolor{gray!10}53.0 \\
    Pick Diverse Bottles & 27.0 & 6.0 & 30.0 & 7.0 & \cellcolor{gray!10}42.0 & \cellcolor{gray!10}8.0 & 51.0 & 40.0 & \cellcolor{gray!10}63.0 & \cellcolor{gray!10}30.0 \\
    Pick Dual Bottles & 57.0 & 12.0 & 43.0 & 13.0 & \cellcolor{gray!10}63.0 & \cellcolor{gray!10}8.0 & 54.0 & 49.0 & \cellcolor{gray!10}69.0 & \cellcolor{gray!10}42.0 \\
    Place A2B Left & 31.0 & 1.0 & 20.0 & 0.0 & \cellcolor{gray!10}62.0 & \cellcolor{gray!10}2.0 & 66.0 & 26.0 & \cellcolor{gray!10}62.0 & \cellcolor{gray!10}39.0 \\
    Place A2B Right & 27.0 & 6.0 & 22.0 & 2.0 & \cellcolor{gray!10}44.0 & \cellcolor{gray!10}3.0 & 52.0 & 17.0 & \cellcolor{gray!10}59.0 & \cellcolor{gray!10}35.0 \\
    Place Bread Basket & 17.0 & 4.0 & 52.0 & 5.0 & \cellcolor{gray!10}61.0 & \cellcolor{gray!10}2.0 & 41.0 & 33.0 & \cellcolor{gray!10}57.0 & \cellcolor{gray!10}46.0 \\
    Place Bread Skillet & 23.0 & 1.0 & 56.0 & 2.0 & \cellcolor{gray!10}52.0 & \cellcolor{gray!10}2.0 & 36.0 & 21.0 & \cellcolor{gray!10}52.0 & \cellcolor{gray!10}29.0 \\
    Place Burger Fries & 80.0 & 4.0 & 96.0 & 6.0 & \cellcolor{gray!10}97.0 & \cellcolor{gray!10}4.0 & 68.0 & 47.0 & \cellcolor{gray!10}85.0 & \cellcolor{gray!10}77.0 \\
    Place Can Basket & 41.0 & 5.0 & 63.0 & 0.0 & \cellcolor{gray!10}68.0 & \cellcolor{gray!10}1.0 & 52.0 & 3.0 & \cellcolor{gray!10}62.0 & \cellcolor{gray!10}12.0 \\
    Place Cans Plasticbox & 34.0 & 2.0 & 81.0 & 2.0 & \cellcolor{gray!10}87.0 & \cellcolor{gray!10}3.0 & 28.0 & 32.0 & \cellcolor{gray!10}19.0 & \cellcolor{gray!10}33.0 \\
    Place Container Plate & 88.0 & 45.0 & 99.0 & 3.0 & \cellcolor{gray!10}94.0 & \cellcolor{gray!10}2.0 & 95.0 & 50.0 & \cellcolor{gray!10}94.0 & \cellcolor{gray!10}76.0 \\
    Place Dual Shoes & 15.0 & 0.0 & 28.0 & 1.0 & \cellcolor{gray!10}2.0 & \cellcolor{gray!10}0.0 & 33.0 & 10.0 & \cellcolor{gray!10}23.0 & \cellcolor{gray!10}7.0 \\
    Place Empty Cup & 37.0 & 11.0 & 72.0 & 4.0 & \cellcolor{gray!10}66.0 & \cellcolor{gray!10}0.0 & 80.0 & 56.0 & \cellcolor{gray!10}89.0 & \cellcolor{gray!10}83.0 \\
    Place Fan & 20.0 & 10.0 & 28.0 & 0.0 & \cellcolor{gray!10}20.0 & \cellcolor{gray!10}0.0 & 57.0 & 24.0 & \cellcolor{gray!10}44.0 & \cellcolor{gray!10}25.0 \\
    Place Mouse Pad & 7.0 & 1.0 & 9.0 & 0.0 & \cellcolor{gray!10}14.0 & \cellcolor{gray!10}0.0 & 24.0 & 23.0 & \cellcolor{gray!10}31.0 & \cellcolor{gray!10}25.0 \\
    Place Object Basket & 16.0 & 2.0 & 40.0 & 1.0 & \cellcolor{gray!10}46.0 & \cellcolor{gray!10}0.0 & 65.0 & 15.0 & \cellcolor{gray!10}72.0 & \cellcolor{gray!10}19.0 \\
    Place Object Scale & 10.0 & 0.0 & 19.0 & 0.0 & \cellcolor{gray!10}32.0 & \cellcolor{gray!10}1.0 & 54.0 & 28.0 & \cellcolor{gray!10}60.0 & \cellcolor{gray!10}38.0 \\
    Place Object Stand & 36.0 & 11.0 & 48.0 & 2.0 & \cellcolor{gray!10}83.0 & \cellcolor{gray!10}0.0 & 77.0 & 31.0 & \cellcolor{gray!10}80.0 & \cellcolor{gray!10}55.0 \\
    Place Phone Stand & 35.0 & 7.0 & 24.0 & 0.0 & \cellcolor{gray!10}63.0 & \cellcolor{gray!10}1.0 & 34.0 & 35.0 & \cellcolor{gray!10}58.0 & \cellcolor{gray!10}39.0 \\
    Place Shoe & 28.0 & 6.0 & 63.0 & 0.0 & \cellcolor{gray!10}14.0 & \cellcolor{gray!10}1.0 & 58.0 & 45.0 & \cellcolor{gray!10}64.0 & \cellcolor{gray!10}42.0 \\
    Press Stapler & 62.0 & 29.0 & 60.0 & 0.0 & \cellcolor{gray!10}78.0 & \cellcolor{gray!10}11.0 & 66.0 & 50.0 & \cellcolor{gray!10}83.0 & \cellcolor{gray!10}55.0 \\
    Put Bottles Dustbin & 54.0 & 13.0 & -- & 0.0 & \cellcolor{gray!10}9.0 & \cellcolor{gray!10}1.0 & 47.0 & 36.0 & \cellcolor{gray!10}68.0 & \cellcolor{gray!10}42.0 \\
    Put Object Cabinet & 68.0 & 18.0 & 35.0 & 0.0 & \cellcolor{gray!10}34.0 & \cellcolor{gray!10}0.0 & 25.0 & 31.0 & \cellcolor{gray!10}41.0 & \cellcolor{gray!10}28.0 \\
    Rotate QRCode & 68.0 & 15.0 & 50.0 & 0.0 & \cellcolor{gray!10}45.0 & \cellcolor{gray!10}0.0 & 64.0 & 23.0 & \cellcolor{gray!10}71.0 & \cellcolor{gray!10}23.0 \\
    Scan Object & 18.0 & 1.0 & 13.0 & 1.0 & \cellcolor{gray!10}30.0 & \cellcolor{gray!10}1.0 & 16.0 & 14.0 & \cellcolor{gray!10}32.0 & \cellcolor{gray!10}18.0 \\
    Shake Bottle & 99.0 & 51.0 & 98.0 & 11.0 & \cellcolor{gray!10}100.0 & \cellcolor{gray!10}23.0 & 95.0 & 87.0 & \cellcolor{gray!10}100.0 & \cellcolor{gray!10}98.0 \\
    Shake Bottle Horizontally & 97.0 & 60.0 & 98.0 & 10.0 & \cellcolor{gray!10}100.0 & \cellcolor{gray!10}36.0 & 96.0 & 94.0 & \cellcolor{gray!10}99.0 & \cellcolor{gray!10}99.0 \\
    Stack Blocks Three & 17.0 & 0.0 & 41.0 & 0.0 & \cellcolor{gray!10}13.0 & \cellcolor{gray!10}0.0 & 37.0 & 0.0 & \cellcolor{gray!10}59.0 & \cellcolor{gray!10}26.0 \\
    Stack Blocks Two & 42.0 & 1.0 & 83.0 & 0.0 & \cellcolor{gray!10}67.0 & \cellcolor{gray!10}1.0 & 59.0 & 15.0 & \cellcolor{gray!10}83.0 & \cellcolor{gray!10}43.0 \\
    Stack Bowls Three & 66.0 & 24.0 & 62.0 & 0.0 & \cellcolor{gray!10}32.0 & \cellcolor{gray!10}0.0 & 53.0 & 44.0 & \cellcolor{gray!10}78.0 & \cellcolor{gray!10}52.0 \\
    Stack Bowls Two & 91.0 & 41.0 & 90.0 & 1.0 & \cellcolor{gray!10}67.0 & \cellcolor{gray!10}0.0 & 92.0 & 66.0 & \cellcolor{gray!10}99.0 & \cellcolor{gray!10}75.0 \\
    Stamp Seal & 3.0 & 4.0 & 27.0 & 0.0 & \cellcolor{gray!10}25.0 & \cellcolor{gray!10}0.0 & 25.0 & 11.0 & \cellcolor{gray!10}51.0 & \cellcolor{gray!10}31.0 \\
    Turn Switch & 27.0 & 23.0 & 26.0 & 0.0 & \cellcolor{gray!10}25.0 & \cellcolor{gray!10}6.0 & 24.0 & 31.0 & \cellcolor{gray!10}39.0 & \cellcolor{gray!10}15.0 \\
    \midrule
    \textbf{Average} & 46.4 & 16.3 & 50.4 & 2.2 & \cellcolor{gray!10}54.6 & \cellcolor{gray!10}4.0 & 54.3 & 33.2 & \cellcolor{gray!10}\textbf{65.5} & \cellcolor{gray!10}\textbf{43.1} \\
    \bottomrule
  \end{tabular}}
\end{table}

\section{Future prediction examples}
\label{sec:appendix_future_prediction}

\begin{figure}[p]
  \centering
  \includegraphics[width=\linewidth]{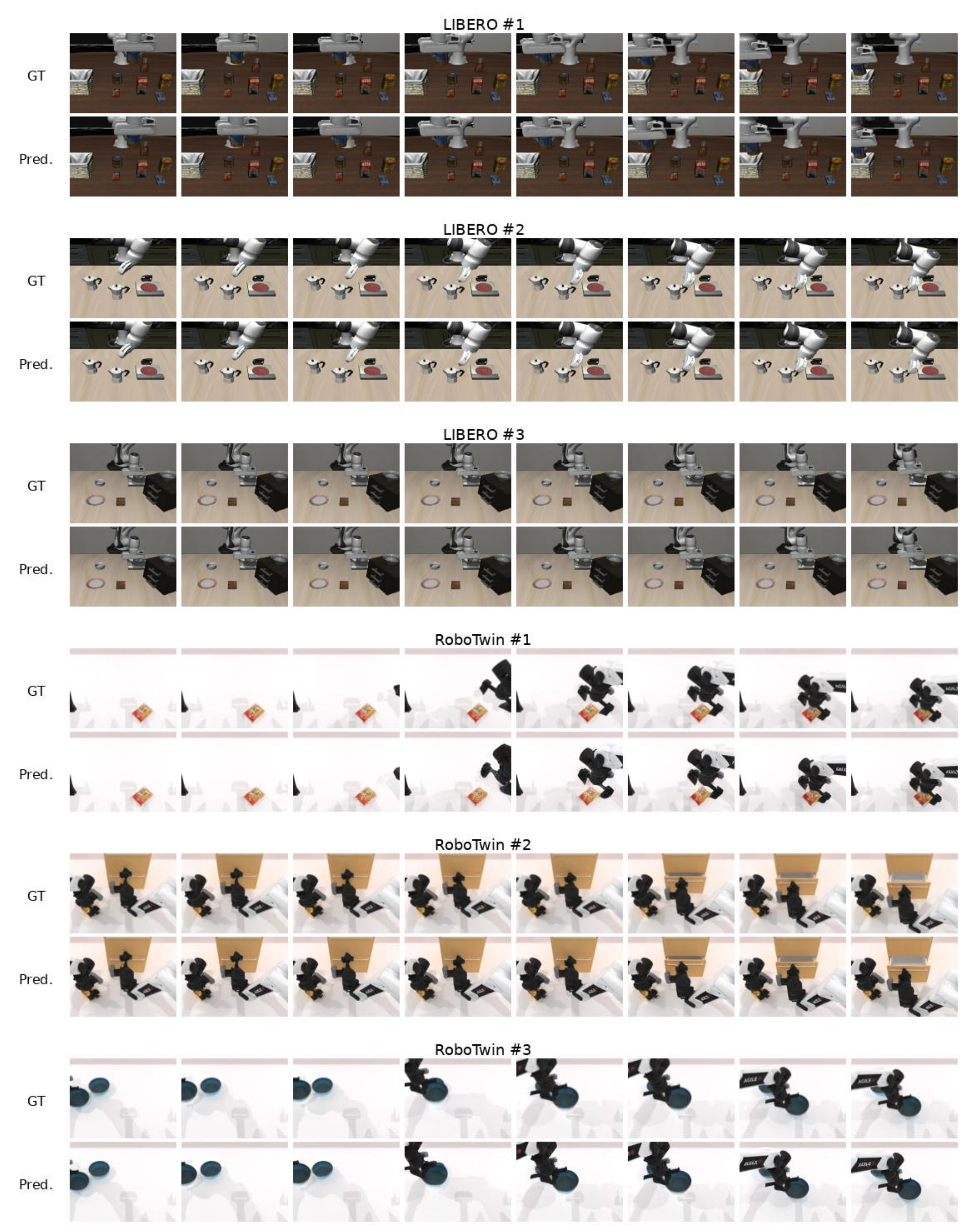}
  \caption{Future visual prediction examples on simulation benchmarks (LIBERO and RoboTwin~2.0). Each example contains a ground-truth row and a prediction row decoded from the future visual latent predicted using the \method{} action latent.}
  \label{fig:appendix_pred_sim}
\end{figure}

\begin{figure}[p]
  \centering
  \includegraphics[width=\linewidth]{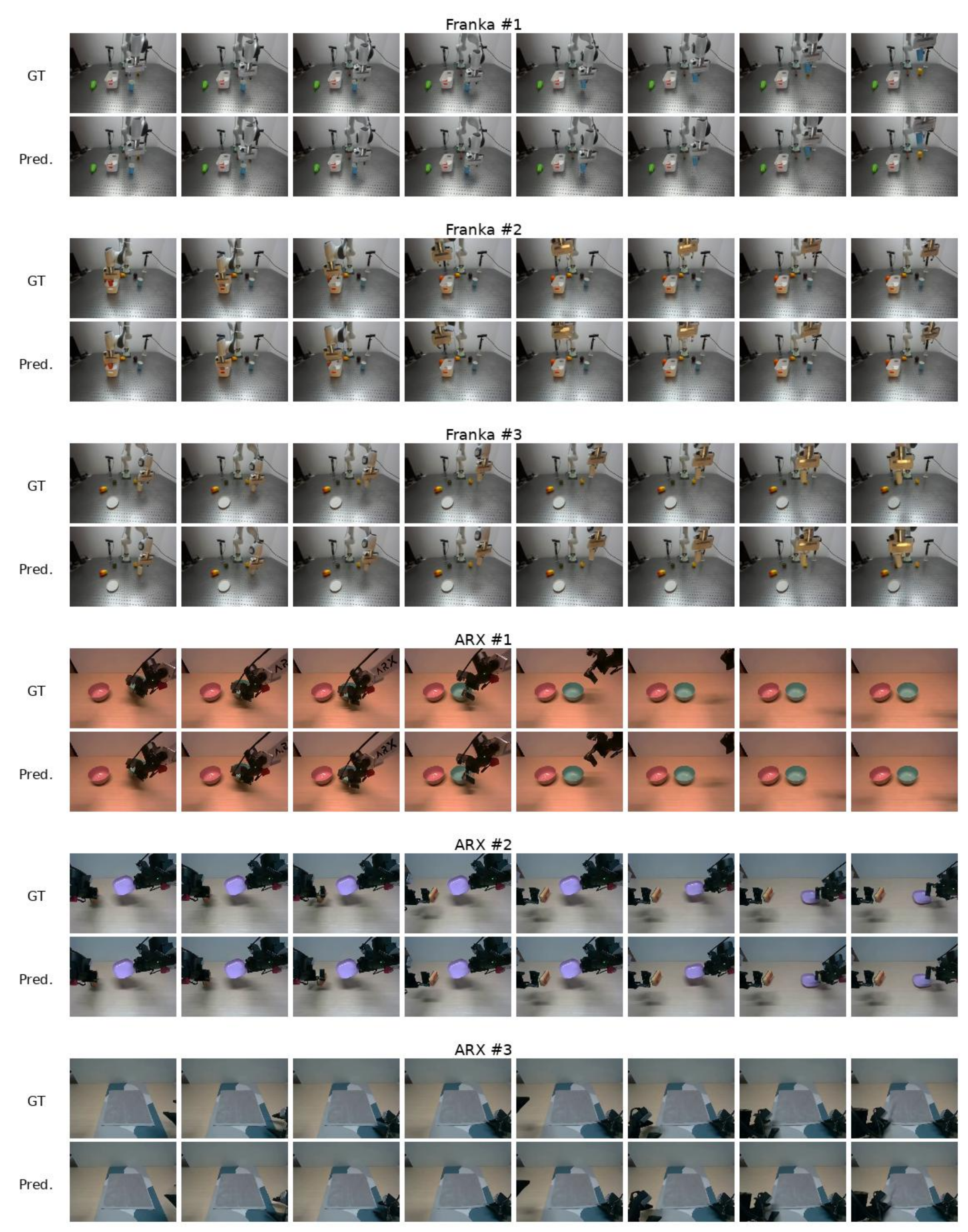}
  \caption{Future visual prediction examples on real-robot datasets (Franka and ARX). Each example contains a ground-truth row and a prediction row decoded from the future visual latent predicted using the \method{} action latent.}
  \label{fig:appendix_pred_real}
\end{figure}

Figures~\ref{fig:appendix_pred_sim} and~\ref{fig:appendix_pred_real} show additional future visual prediction examples from the learned action latent across simulation and real-robot data. For each example, the top row shows ground-truth future frames from the trajectory, and the bottom row shows frames decoded from the predicted future video latent conditioned on the \method{} action latent. These visualizations are used only as qualitative analysis of the latent space; the video VAE, future-latent predictor, and video decoder are not part of the deployed policy.

\end{document}